%% file: main.tex
\documentclass[10pt,twocolumn,letterpaper]{article}

\usepackage{cvpr}              

\input{preamble}

\definecolor{cvprblue}{rgb}{0.21,0.49,0.74}
\usepackage[pagebackref,breaklinks,colorlinks,allcolors=cvprblue]{hyperref}
\usepackage{cuted}   

\newcommand\blfootnote[1]{%
\begingroup
\renewcommand\thefootnote{}\footnote{#1}%
\addtocounter{footnote}{-1}%
\endgroup
}

\def\paperID{****} 
\def\confName{CVPR}
\def\confYear{2026}

\title{FreqEdit: Preserving High-Frequency Features for Robust \\Multi-Turn Image Editing}

\author{
Yucheng Liao$^{1,*}$\quad 
Jiajun Liang$^{1,*}$\quad 
Kaiqian Cui$^{1,*}$\\
Baoquan Zhao$^{1}$\quad 
Haoran Xie$^{2}$\quad 
Wei Liu$^{3}$\quad 
Qing Li$^{4}$\quad 
Xudong Mao$^{1,\dagger}$\\
{$^1$School of Artificial Intelligence, Sun Yat-sen University}\ \ {$^2$Lingnan University}\\{$^3$Video Rebirth}\ \ 
{$^4$The Hong Kong Polytechnic University }\\
{\href{https://freqedit.github.io/}{https://freqedit.github.io/ \vspace{-15pt}}}
}

\begin{document}
\input{fig_tex/teaser}

\blfootnote{\textsuperscript{*} Equal contribution, 
\textsuperscript{$\dagger$} Corresponding author.}

\input{0_abstract}
\input{1_intro}

\input{2_related}
\input{3_preliminary}
\input{4_method}

\input{5_experiment}

\input{6_conclusion}
{
    \small
    \bibliographystyle{ieeenat_fullname}
    \bibliography{main}
}

\input{7_appendix}

\end{document}

%% file: preamble.tex
\usepackage{booktabs}   
\usepackage{makecell}  
\usepackage{multirow}   
\usepackage{stfloats} 

\usepackage[table]{xcolor}
\definecolor{kontextblue}{RGB}{230,245,255} 
\definecolor{qwenred}{RGB}{255,235,235}      
\definecolor{kontextblueDeep}{RGB}{180,215,255} 
\definecolor{qwenredDeep}{RGB}{255,200,200}  

\usepackage{tcolorbox}
\usepackage{xcolor}

\newcommand{\apptocsection}[3]{
  \noindent\textbf{\hyperref[#3]{#1\ #2}}
  \dotfill\hyperref[#3]{\textbf{\pageref{#3}}}\par
}

\newcommand{\apptocsubsection}[3]{
  \noindent\hspace*{1.8em}\hyperref[#3]{#1\ #2}%
  \dotfill\hyperref[#3]{\pageref{#3}}\par
}

%% file: fig_tex/teaser.tex
\twocolumn[{
\maketitle
\begin{center}
    \centering
    \vspace{-6pt}
    \includegraphics[width=1.0\textwidth]{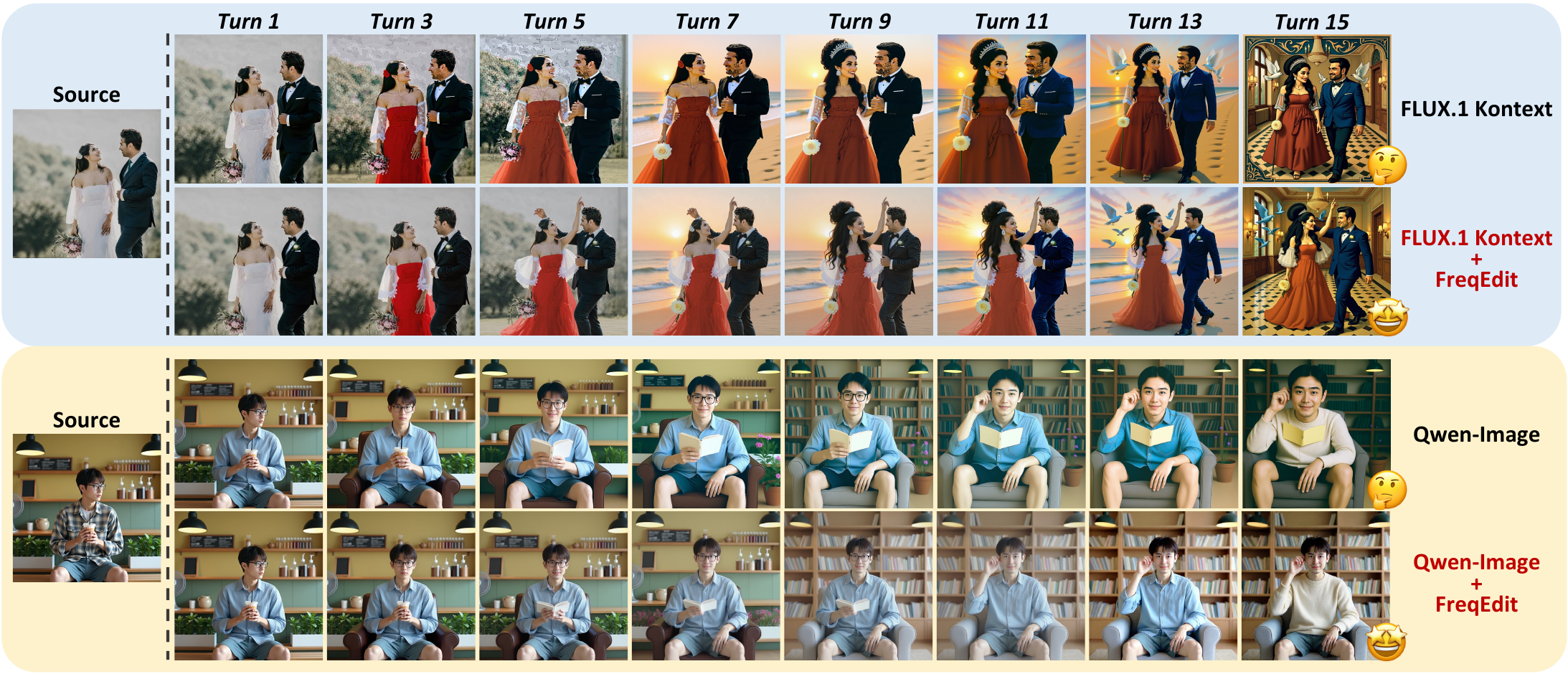}
    \captionof{figure}{\textbf{FreqEdit enables consistent multi-turn image editing.} Base models (FLUX.1 Kontext and Qwen-Image) exhibit progressive quality deterioration during iterative editing, including body deformations, edge over-sharpening, and texture collapse. FreqEdit addresses these limitations through strategic high-frequency reinforcement.
    }
    \vspace{-8pt}
    \label{fig:teaser}
\end{center}
}]

%% file: 0_abstract.tex
\begin{abstract}
Instruction-based image editing through natural language has emerged as a powerful paradigm for intuitive visual manipulation. While recent models achieve impressive results on single edits, they suffer from severe quality degradation under multi-turn editing. Through systematic analysis, we identify progressive loss of high-frequency information as the primary cause of this quality degradation. We present FreqEdit, a training-free framework that enables stable editing across 10+ consecutive iterations. Our approach comprises three synergistic components: (1) high-frequency feature injection from reference velocity fields to preserve fine-grained details, (2) an adaptive injection strategy that spatially modulates injection strength for precise region-specific control, and (3) a path compensation mechanism that periodically recalibrates the editing trajectory to prevent over-constraint.  Extensive experiments demonstrate that FreqEdit achieves superior performance in both identity preservation and instruction following compared to seven state-of-the-art baselines.
\vspace{-8pt}
\end{abstract}

%% file: 1_intro.tex
\section{Introduction}
\label{sec:intro}
\vspace{-4pt}
Recent advances in instruction-based image editing~\cite{brooks2022instructpix2pix,Zhang2023MagicBrush, zhang2025context, 
SuperEdit, Smartedit, fu2024mgie, ge2024seeddataedittechnicalreporthybrid,  
Geng23instructdiff, OmniGen, ACE,  guo2023focusinstructionfinegrainedmultiinstruction,
labs2025flux1kontextflowmatching, wu2025qwenimagetechnicalreport} have unlocked unprecedented creative accessibility, allowing users to perform sophisticated visual manipulations through natural language commands. While recent methods have achieved impressive results on single-turn editing tasks, they struggle to support the iterative, multi-turn workflows that are fundamental to real-world creative processes. For instance, professional photographers progressively refine portraits through dozens of sequential adjustments: first correcting exposure and lighting, then fine-tuning skin tones, modifying hair color, adding accessories, and finally applying stylistic filters. Each step builds upon the previous modifications, requiring the model to maintain coherent editing history while enabling precise control. Although several recent methods~\cite{EmuEdit, MTC, qu2025vincie} have attempted to enhance multi-turn editing capabilities, they still face fundamental limitations in preserving character consistency and maintaining editing precision.

Our empirical investigation of several state-of-the-art instruction-based editing models, including FLUX.1 Kontext~\cite{labs2025flux1kontextflowmatching} and Qwen-Image~\cite{wu2025qwenimagetechnicalreport}, reveals that even these sophisticated models maintain reliable editing quality for approximately five sequential edits before experiencing severe degradation. As illustrated in Figure~\ref{fig:teaser}, extended editing sequences spanning ten or more iterations lead to catastrophic quality deterioration. Through extensive analysis, we identify three pervasive failure modes: 1) \textit{subject deformation}, where the primary subject's geometric structure and appearance progressively deviate from the original identity, 2) \textit{edge over-sharpening}, where boundaries become artificially enhanced, and 3) \textit{texture collapse}, where fine-grained details (e.g., skin pores) degrade into overly smooth surfaces or artifacts.

What causes this systematic degradation? We hypothesize that the accumulated errors in high-frequency features across editing iterations are the root cause. To validate this, we conduct controlled ablation experiments by artificially manipulating high-frequency components in source images through spatial filtering. Specifically, we amplify high-frequency edges via unsharp masking \cite{gonzalez2008digital} or suppress high-frequency textures via bilateral filtering \cite{bilateralfiltering}. As shown in Figure~\ref{fig:control}, both interventions dramatically accelerate degradation, with subject deformation emerging as early as the third editing turn. This confirms that high-frequency features serve as critical identity anchors. The underlying degradation mechanism can be understood as follows: high-frequency components encode essential identity-specific structures and fine-grained details. When this information degrades across iterations, the generative model increasingly relies on its learned priors, regressing toward canonical representations prevalent in training data (e.g., frontal poses, average face sizes).

Motivated by this insight, we introduce \textit{FreqEdit}, a training-free framework that maintains high visual fidelity and instruction-following capability across extended multi-turn editing sessions. Our core strategy is to selectively reinforce high-frequency information during early denoising steps, when these components are most vulnerable to degradation. This vulnerability arises because early denoising steps primarily establish low-frequency global structure, making high-frequency details susceptible to suppression. Specifically, we first construct a reference velocity field from the context image (i.e., the input image for the current editing turn) containing rich high-frequency details, and then inject these high-frequency components into the editing velocity field to counteract progressive degradation. However, naively applying uniform injection would either overly constrain edited regions or compromise the integrity of unedited areas. We therefore propose an adaptive injection strategy that spatially modulates the reference strength based on automatically predicted editing masks, ensuring that unedited regions maintain their original fidelity while edited regions retain sufficient flexibility for meaningful transformations. Furthermore, continuous high-frequency injection can overly restrict the generation process, potentially leading to incomplete or suboptimal edits. To address this, we introduce a path compensation mechanism that periodically recalibrates the editing trajectory, dynamically steering it back toward the intended editing direction.

\input{fig_tex/control}

%% file: fig_tex/control.tex
\begin{figure}[tb]
    \centering
    \vspace{-4pt}
    \includegraphics[width=\columnwidth]{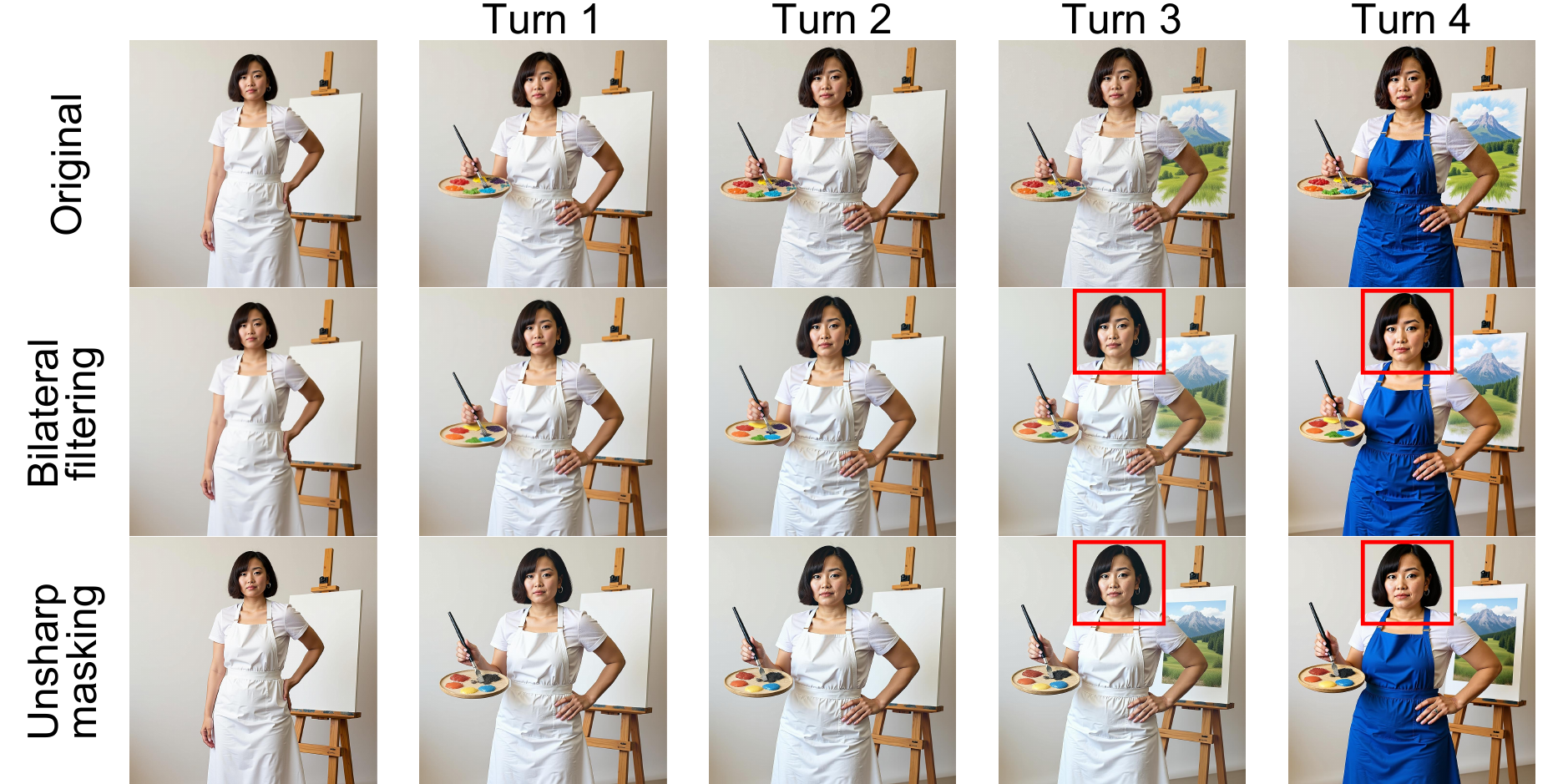}
    \caption{
    Bilateral filtering and unsharp masking are each applied once to the source image before editing. Both perturbations accelerate quality degradation, causing subject deformation as early as Turn~3 (red boxes). This confirms that high-frequency integrity is critical for maintaining subject identity across editing turns.
    }
    \label{fig:control}
    \vspace{-12pt}
\end{figure}

%% file: 2_related.tex
\section{Related Work}

\noindent\textbf{Image Editing.}
Building on advances in diffusion models~\cite{LDM, sdxl} and flow models~\cite{liu2022flow, lipman2023flowmatchinggenerativemodeling, flux2024, sd3}, image editing has witnessed significant progress across multiple paradigms. Attention manipulation approaches~\cite{hertz2022prompt, Tumanyan_2023_CVPR, li2023stylediffusion, cao_2023_masactrl, chefer2023attendandexcite} enable localized modifications while preserving image structure. Mask-based methods~\cite{couairon2022diffeditdiffusionbasedsemanticimage, Avrahami_2022, lugmayr2022repaintinpaintingusingdenoising, huang2023region} and inversion techniques~\cite{DDIM, mokady2022null, miyake2024negativepromptinversionfastimage, Avrahami_2025_CVPR,deng2024fireflowfastinversionrectified, wang2024taming, rout2025semantic} further improve editing control through explicit spatial guidance and latent space manipulation.
Frequency-based methods~\cite{wu2024freediffprogressivefrequencytruncation, fds,gao2024frequency} enable detail manipulation and style editing via frequency component modulation, yet their applicability to multi-turn editing remains largely unexplored.

\vspace{3pt}
\noindent\textbf{Instruction-based Image Editing.}
Within the landscape of image editing methods, instruction-based editing has emerged as a particularly intuitive approach. InstructPix2Pix~\cite{
brooks2022instructpix2pix} pioneered this paradigm by leveraging synthetic 
paired data and model fine-tuning. Subsequent work has focused on 
improved training strategies~\cite{Zhang2023MagicBrush, zhang2025context, 
SuperEdit,simsar2025uip2punsupervisedinstructionbasedimage,zhang2024hiveharnessinghumanfeedback,sun2023imagebrushlearningvisualincontext}, enhanced multimodal understanding~\cite{Smartedit,he2024freeeditmaskfreereferencebasedimage, 
Geng23instructdiff, EmuEdit, fu2024mgie}, and unified architectures~\cite{
OmniGen, ACE}. More recently, context-based flow models~\cite{
step1x-edit, labs2025flux1kontextflowmatching, wu2025qwenimagetechnicalreport} 
have demonstrated the effectiveness of in-context learning for coherent 
edits. Despite these advances, maintaining consistency across multiple 
sequential edits remains an open challenge.

\vspace{3pt}
\noindent\textbf{Multi-turn Image Editing.}
Several prior works attempt to address the challenge of sequential consistency in multi-turn editing. Emu Edit~\cite{EmuEdit} mitigates error accumulation via corrective processing and per-pixel thresholding. MTC~\cite{MTC} employs trajectory control and adaptive attention guidance to preserve coherence. VINCIE~\cite{qu2025vincie} trains a block-causal diffusion transformer on video data as interleaved multi-modal sequences. 
Although these methods demonstrate improved multi-turn capabilities, they lack a systematic understanding of the mechanisms causing cumulative degradation, which limits their ability to fundamentally prevent this phenomenon while maintaining editing quality.

%% file: 3_preliminary.tex
\section{Preliminary}
\label{sec:preliminary}
\noindent\textbf
\noindent\textbf{Rectified Flow}~\cite{liu2022flow, lipman2023flowmatchinggenerativemodeling} establishes a continuous transport between noise $\pi_1 = \mathcal{N}(0, \mathbf{I})$ and data distribution $\pi_0$ via an ODE. The model learns a velocity field $v_\theta(Z_t, t, \mathbf{c})$ where $t \in [0, 1]$ and $\mathbf{c}$ denotes conditioning. Given training pairs $X_0 \sim \pi_0$ and $X_1 \sim \pi_1$, rectified flow constructs linear interpolations $X_t = (1-t)X_0 + tX_1$ and minimizes:
\begin{equation}
\mathcal{L} = \mathbb{E}_{t, X_0, X_1} \left[\left\| v_\theta(X_t, t, \mathbf{c}) - (X_1 - X_0) \right\|^2_2 \right].
\end{equation}
During inference, generation follows the ODE $\frac{dZ_t}{dt} = v_\theta(Z_t, t, \mathbf{c})$ starting from $Z_1 \sim \mathcal{N}(0, \mathbf{I})$, discretized via Euler steps:
\begin{equation}
\label{eq:euler}
Z_{t_{i-1}} = Z_{t_i} + (t_{i-1} - t_i) \cdot v_\theta(Z_{t_i}, t_i, \mathbf{c}),
\end{equation}
producing the final sample at $t=0$.

%% file: 4_method.tex
\input{fig_tex/framework}
\section{Method}
\label{sec:method}

\subsection{Overview}
\label{sec:overview}
Iterative image editing faces a fundamental challenge: the progressive degradation of high-frequency information across editing turns. This degradation manifests as subject deformation, edge over-sharpening, and texture collapse, severely compromising the visual fidelity of edited images. The root cause lies in an inherent characteristic of the denoising process: during early timesteps, when the noisy image remains close to Gaussian noise, the predicted velocity field lacks sufficient information to accurately recover high-frequency components. When this information degrades across iterations, the generative model increasingly relies on its learned priors, regressing toward canonical representations prevalent in training data. 

Our key insight is that the context image (i.e., the input image for the current editing turn) contains rich high-frequency information that can be leveraged to compensate for this degradation. However, naively injecting these components would conflict with the editing objective, potentially suppressing desired semantic transformations. We therefore propose a principled wavelet-based framework that strategically integrates high-frequency information while respecting the editing instruction.

As illustrated in Figure~\ref{fig:framework}, our approach begins by constructing a reference velocity field from the context image and injecting its high-frequency components into the editing velocity field (Section~\ref{sec:high_freq_injection}). To address the challenge that uniform injection strength can degrade editing quality in regions undergoing semantic modification, we introduce a spatially-adaptive injection mechanism (Section~\ref{sec:adaptive}). To eliminate the ghosting artifacts where conflicting visual elements from the editing and reference velocities manifest simultaneously, we propose a path compensation strategy (Section~\ref{sec:path_compensation}) that periodically compensates the denoising trajectory toward the desired editing direction. Finally, for models exhibiting noise accumulation, we introduce a quality-guided refinement mechanism (Section~\ref{sec:quality_preservation}) that blends the editing velocity with an auxiliary velocity from the original image during final denoising steps.

\subsection{Wavelet-based Feature Injection}
\label{sec:high_freq_injection}

At the $k$-th editing turn, given the context image $X^{[k]}$, the noisy image $Z_{t_i}$ at timestep $t_i$, and the text instruction ${p}^{[k]}$, the editing velocity field is predicted as:
\begin{equation}
v^{\text{edit}}_{t_i} = v_\theta(Z_{t_i}, t_i, X^{[k]}, {p}^{[k]}).
\end{equation}
To address high-frequency component degradation, we propose leveraging the rich high-frequency information preserved in the context image. Specifically, we first construct a reference velocity field from the context image and then strategically inject its high-frequency components into the editing velocity field.

\vspace{3pt}
\noindent\textbf{Reference Velocity Construction.}
For simplicity, we omit turn $k$ and let $Z^{\text{ref}}_0 = X^{[k]}$. We construct the reference velocity field from the context image as:
\begin{equation}
v^{\text{ref}}_{t_i} = \frac{Z^{\text{ref}}_0 - Z_{t_i}}{t_0 - t_i},
\label{eq:v_ref}
\end{equation}
where $t_0$ represents the terminal timestep of the denoising trajectory. This formulation is derived from the Euler discretization (Eq.~\ref{eq:euler}), which can be expressed as ${v} = \frac{Z_{t_{i-1}} - Z_{t_i}}{t_{i-1} - t_i}$.
The complete derivation is provided in Appendix~\ref{sec:appendix_derivation}. Intuitively, as illustrated in Figure~\ref{fig:framework}(A), $v^{\text{ref}}_{t_i}$ represents the ``average velocity'' from the current position $Z_{t_i}$ toward the context image $Z^{\text{ref}}_0$, thereby constructing a straight-line trajectory in the latent space that preserves the high-frequency characteristics of the context image.

\vspace{3pt}
\noindent\textbf{High-Frequency Feature Extraction.}
To selectively extract high-frequency components from the reference velocity, we employ a 2-level discrete wavelet transform (DWT)~\cite{daubechies1992ten}. The multi-level decomposition enables the capture of high-frequency information across multiple scales: fine-grained details such as skin pores and sharp edge structures at the first level, and coarser textural patterns such as fabric weaves at the second level.

Formally, we apply the 2-level DWT to both the reference velocity ${v}^{\text{ref}}$ and the editing velocity ${v}^{\text{edit}}$ (omitting the timestep $t_i$ for brevity):
\begin{align}
&\text{DWT}({v}^{\text{ref}}) = \{\mathbf{LL}^{(2)}_{\text{ref}}, \mathbf{D}^{(2)}_{\text{ref}}, \mathbf{D}^{(1)}_{\text{ref}}\},\\
&\text{DWT}({v}^{\text{edit}}) = \{\mathbf{LL}^{(2)}_{\text{edit}}, \mathbf{D}^{(2)}_{\text{edit}}, \mathbf{D}^{(1)}_{\text{edit}}\},
\end{align}
where $\mathbf{LL}^{(2)}$ denotes the second-level low-frequency approximation coefficients, and $\mathbf{D}^{(\ell)} = \{\mathbf{LH}^{(\ell)}, \mathbf{HL}^{(\ell)},$ $\mathbf{HH}^{(\ell)}\}$ for $\ell \in \{1, 2\}$ represents the high-frequency detail coefficients at level $\ell$ in the vertical, horizontal, and diagonal directions, respectively. It is important to note that only the high-frequency components $\{\mathbf{D}^{(2)}_{\text{ref}}, \mathbf{D}^{(1)}_{\text{ref}}\}$ from the reference velocity are injected into the editing velocity, as the low-frequency components encode the global structure and semantic layout that should be controlled by the editing instruction, while the high-frequency components capture texture patterns and edge sharpness that are relatively content-agnostic.

\vspace{3pt}
\noindent\textbf{High-Frequency Feature Injection.}
Based on the extracted high-frequency coefficients from the reference velocity, we now inject them into the editing velocity to compensate for the progressive loss of high-frequency information. Inspired by classifier-free guidance~\cite{cfg}, we propose an injection mechanism that performs linear extrapolation in the frequency domain by scaling the difference between reference and editing coefficients. Specifically, for each level $\ell$ and each component in $\mathbf{D}^{(\ell)}$, the corrected high-frequency coefficients are computed as:
\begin{equation}
\tilde{\mathbf{D}}^{(\ell)} = \mathbf{D}^{(\ell)}_{\text{edit}} + \alpha (\mathbf{D}^{(\ell)}_{\text{ref}} - \mathbf{D}^{(\ell)}_{\text{edit}}),
\label{eq:injection}
\end{equation}
where $\alpha$ controls the injection strength.

Finally, the corrected velocity field is reconstructed via the inverse DWT:

\begin{equation}
{v}^{\text{corr}} = \text{IDWT}(\mathbf{LL}^{(2)}_{\text{edit}}, \tilde{\mathbf{D}}^{(2)}, \tilde{\mathbf{D}}^{(1)}),
\label{eq:IDWT}
\end{equation}
where $\mathbf{LL}^{(2)}_{\text{edit}}$ is the low-frequency coefficient from the editing velocity. This corrected velocity field ${v}^{\text{corr}}$ is then used for the subsequent denoising step.

\subsection{Adaptive Injection Strategy}
\label{sec:adaptive}
While Eq.~\ref{eq:injection} successfully transfers high-frequency components from the reference velocity, applying uniform injection strength $\alpha$ across all spatial locations can degrade editing quality. Regions requiring substantial semantic modifications may suffer from overly aggressive injection, which suppresses desired transformations and causes unintended preservation of reference image characteristics. 

We address this challenge through a spatially-adaptive injection mechanism that modulates injection strength based on the semantic correspondence between editing and reference velocities. Our key insight is that spatial locations with minimal velocity divergence indicate semantically consistent regions that should receive stronger high-frequency injection to preserve details, while locations with substantial divergence indicate areas undergoing semantic editing that require attenuated injection to accommodate the transformation.

We first quantify the spatial divergence between ${v}^{\text{edit}}$ and ${v}^{\text{ref}}$ by computing the $L_2$ norm of their difference across the channel dimension:
\begin{equation}
\mathbf{M} = \|{v}^{\text{edit}} - {v}^{\text{ref}}\|_2.
\end{equation}
This produces a 2D difference map that captures regions with varying degrees of semantic correspondence.

To convert the difference map into injection strengths, we normalize $\mathbf{M}$ to $[0,1]$ and invert it such that smaller differences yield higher injection values:
\begin{equation}
\tilde{\mathbf{M}} = 1 - \frac{\mathbf{M} - \min(\mathbf{M})}{\max(\mathbf{M}) - \min(\mathbf{M})}.
\end{equation}
We then apply exponential scaling to amplify the contrast between preservation regions (high injection) and editing regions (low injection):
\begin{equation}
\alpha = \alpha_0 \left( e^{\gamma \cdot \tilde{\mathbf{M}}} - 1 \right),
\label{eq:exponential_scaling}
\end{equation}
where $\alpha_0$ controls the overall injection magnitude and $\gamma$ governs the sharpness of the transition between preservation and editing regions.

The adaptive injection strength is incorporated into Eq.~\ref{eq:injection} through element-wise modulation:
\begin{equation}
\tilde{\mathbf{D}}^{(\ell)} = \mathbf{D}^{(\ell)}_{\text{edit}} + \boldsymbol{\alpha}^{(\ell)} \odot \left(\mathbf{D}^{(\ell)}_{\text{ref}} - \mathbf{D}^{(\ell)}_{\text{edit}}\right),
\label{eq:adaptive_injection}
\end{equation}
where $\boldsymbol{\alpha}^{(\ell)}$ is the adaptive injection strength map for decomposition level $\ell$, and $\odot$ denotes element-wise multiplication. Finally, we derive ${v}^{\text{corr}}$ following Eq.~\ref{eq:IDWT}.

\input{fig_tex/path_compensation}

\subsection{Path Compensation}
\label{sec:path_compensation}

While the adaptive injection strategy (Section~\ref{sec:adaptive}) spatially modulates injection strength based on semantic correspondence, it reduces rather than eliminates high-frequency injection in edited regions. In scenarios requiring strong injection to prevent subject deformation, residual high-frequency signals in semantically modified regions can still cause ghosting artifacts, where visual elements from both the editing and reference velocity fields manifest simultaneously. For instance, as shown in Figure~\ref{fig:path_compensation}, when the editing instruction modifies the hand positions, the generated image may exhibit inconsistent geometry.

In addition, the required injection strength varies across inputs: some images demand stronger injection to maintain subject fidelity, necessitating a conservatively high global setting that inevitably amplifies residual conflicts in edited regions. This tension between sufficient high-frequency preservation and artifact-free editing motivates our path compensation strategy.

Our core idea is to periodically realign the denoising trajectory toward the desired editing direction after several injection steps. Specifically, as illustrated in Figure~\ref{fig:path_compensation}, we periodically perform $n$ consecutive injection steps using Eq.~\ref{eq:adaptive_injection}. During this process, we track the cumulative trajectory divergence by computing the velocity difference between ${v}^{\text{edit}}$ and ${v}^{\text{corr}}$ at each injection step: 
\begin{equation}
\Delta {v}_{t_i} = {v}^{\text{edit}}_{t_i} - {v}^{\text{corr}}_{t_i}.
\end{equation}
We accumulate this divergence in a trajectory buffer ${B}$, weighted by the timestep interval:
\begin{equation}
{B} \leftarrow {B} + (t_{i-1} - t_i) \cdot \Delta {v}_{t_i}.
\label{eq:trajectory_buffer}
\end{equation}
After $n$ injection steps, we apply path compensation to realign the trajectory with the editing objective:
\begin{equation}
{Z}_{t_{i-n}} \leftarrow {Z}_{t_{i-n}} + {B},
\end{equation}
and reset ${B} = 0$. This compensation operation is performed every $n$ steps throughout the denoising process, as well as at the final injection step to ensure complete trajectory alignment.

The core motivation behind path compensation is to inject sufficient high-frequency features at each denoising step while periodically steering the trajectory toward the target edit semantics by incorporating the accumulated correction $\Delta v$. As illustrated in Figure~\ref{fig:path_compensation}, the path compensation mechanism retraces the current point along the accumulated velocity difference $\Delta {v}_{t_i}$ (depicted as the purple path). Crucially, the actual trajectory, which comprises the orange injection path followed by the purple compensation path, is mathematically equivalent to the blue dashed path governed entirely by ${v}^{\text{edit}}$. This blue dashed path can therefore be interpreted as predicting ${v}^{\text{edit}}$ conditioned on high-frequency information from ${v}^{\text{ref}}$ and performing denoising along this editing velocity.

\subsection{Quality Guidance for Noise Suppression}
\label{sec:quality_preservation}
While our high-frequency injection strategies address the challenges discussed in Section~\ref{sec:overview} (e.g., subject deformation), certain models (e.g., FLUX.1 Kontext) tend to exhibit noise artifacts (see Figure~\ref{fig:ablation}) after multiple editing iterations. This degradation stems from the accumulation of noise introduced in each editing turn, which gradually compromises image quality.

To mitigate this issue, we leverage two key observations: (1) the final steps of the denoising process primarily focus on refining fine details rather than generating semantic content, and (2) the original image (i.e., the input to the first editing turn) contains the highest quality information with minimal noise artifacts. Building on these insights, we introduce a quality-guided refinement mechanism that preserves the visual fidelity of the original image while maintaining editing consistency.

Specifically, during the final denoising steps ($ t_i<\tau_{\text{guide}}$), we blend the editing velocity with an auxiliary velocity constructed from the original image $X^{[1]}$:
\begin{equation}
{v}^{\text{final}}_{t_i} = (1 - \lambda) \cdot {v}^{\text{edit}}_{t_i} + \lambda \cdot {v}_\theta({Z}_{t_i}, t_i, X^{[1]}, {p}_{\text{neutral}}),
\label{eq:quality_guidance}
\end{equation}
where $\lambda \in [0, 1]$ controls the strength of quality guidance from the original image, and ${p}_{\text{neutral}}$ is a neutral prompt (e.g., ``a high-quality picture. '') that avoids introducing new semantic information. The threshold $\tau_{\text{guide}}$ determines when this refinement mechanism is activated.

Note that in our experiments, this refinement strategy was specifically applied to FLUX.1 Kontext, as other models did not exhibit significant noise accumulation issues.

%% file: fig_tex/framework.tex
\begin{figure*}[htb]
    \centering
    \includegraphics[width=\textwidth]{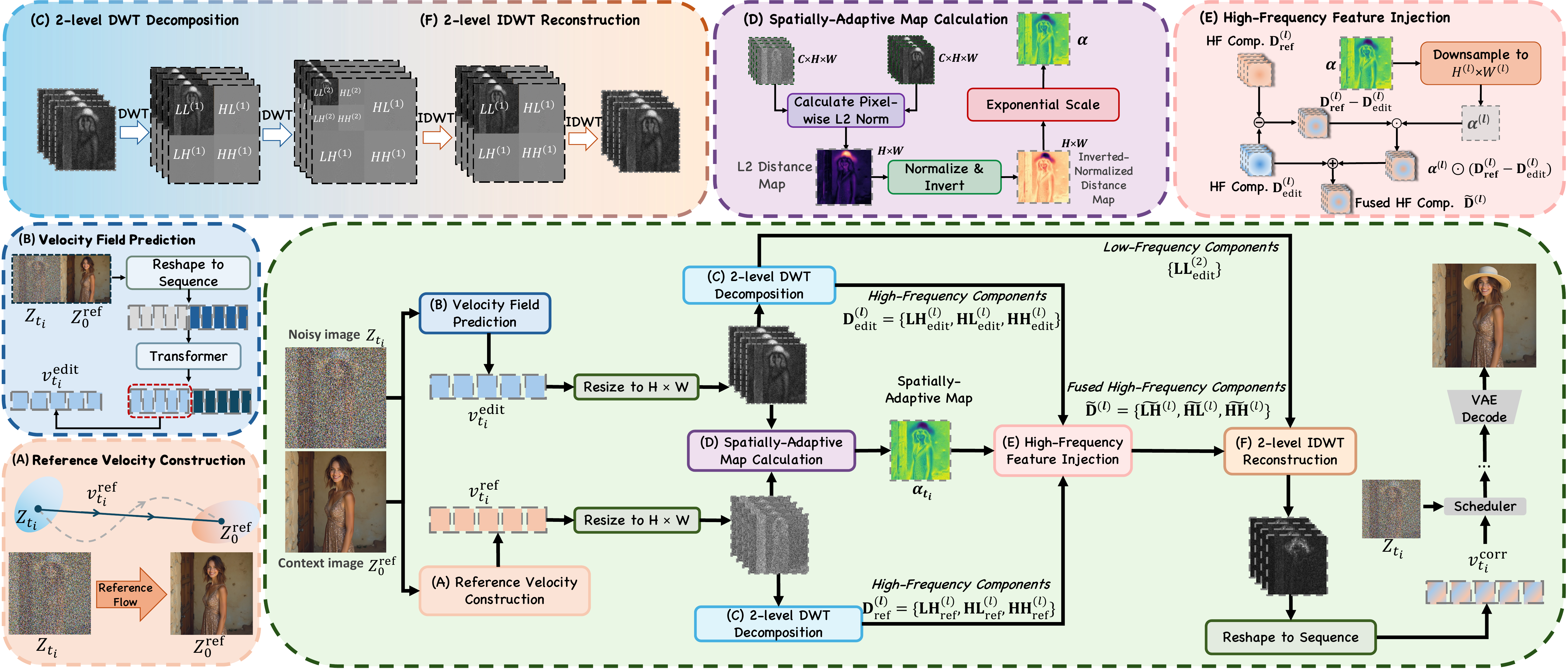}
\caption{\textbf{High-Frequency Feature Injection Pipeline.} (A) We construct the reference velocity $v^{\text{ref}}$ from the context image $Z^{\text{ref}}_0$ using Eq.~\ref{eq:v_ref} (B) The editing velocity $v^{\text{edit}}$ corresponds to the standard prediction of the model. (C) 2-level DWT decomposition extracts multi-scale high-frequency components from both $v^{\text{ref}}$ and $v^{\text{edit}}$. (D) A spatially-adaptive weight map $\boldsymbol{\alpha}$ is computed from the editing mask. (E) High-frequency components are fused via a CFG-style formulation (Eq.~\ref{eq:adaptive_injection}). (F) IDWT reconstruction yields the corrected velocity $v^{\text{corr}}$ by combining fused high-frequency components with low-frequency components from $v^{\text{edit}}$.}
\label{fig:framework}
\vspace{-10pt}
\end{figure*}

%% file: fig_tex/path_compensation.tex
\begin{figure}[htb]
    \centering
    \vspace{-4pt}
    \includegraphics[width=\columnwidth]
    {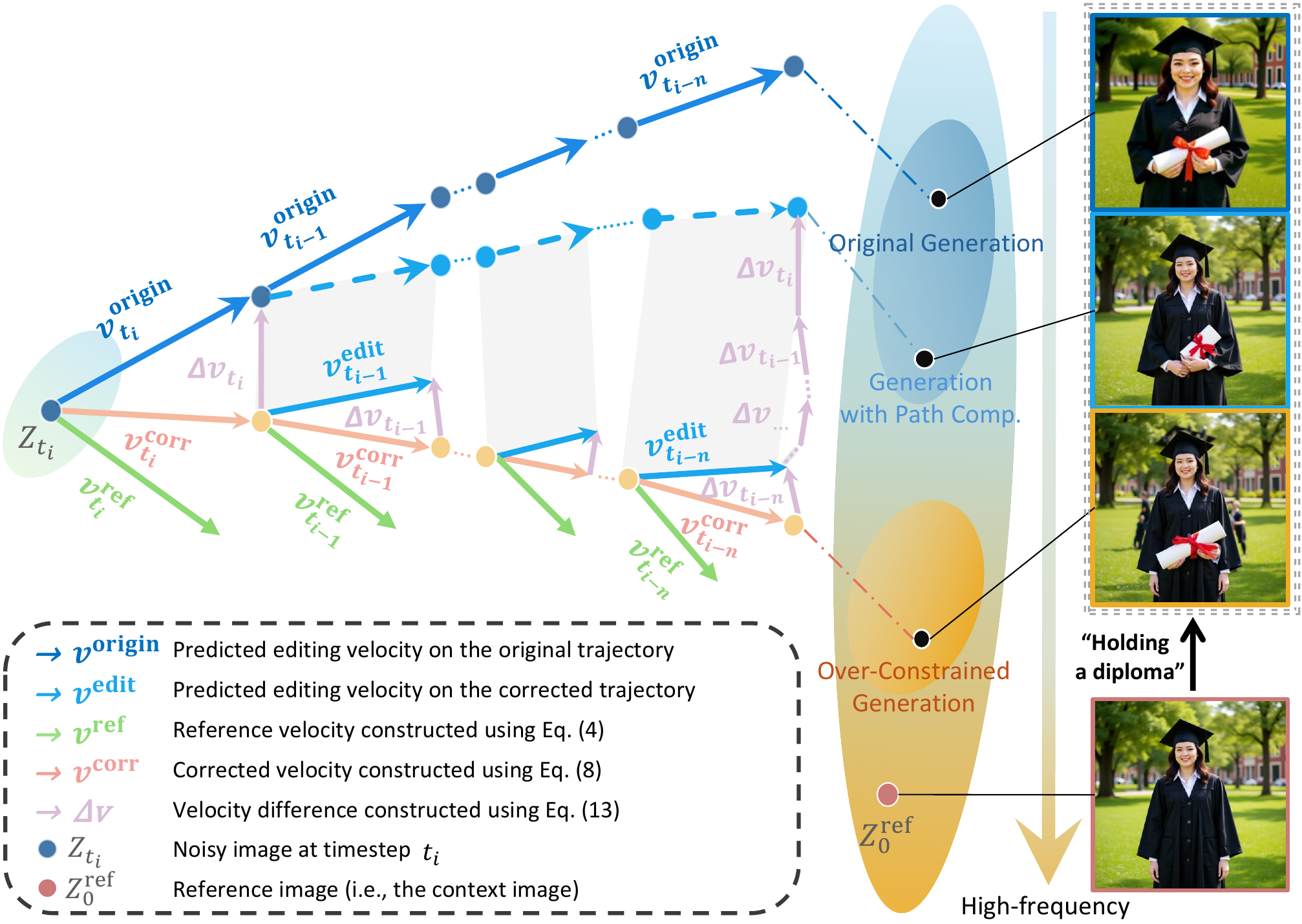}
    \caption{\textbf{Path Compensation Mechanism.} The actual denoising trajectory (orange line ${v}^{\text{corr}}$ to purple line $\Delta v$) is equivalent to the blue dashed trajectory that is entirely governed by the editing velocity ${v}^{\text{edit}}$. This equivalence can be interpreted as predicting $v^{\text{edit}}$ conditioned on high-frequency information from the reference velocity $v^{\text{ref}}$ and  performing denoising along this editing direction.}

    \label{fig:path_compensation}
    \vspace{-9pt}
\end{figure}

%% file: 5_experiment.tex
\section{Experiments}

\input{fig_tex/qualitative}

\subsection{Implementation and Evaluation Setup}

\noindent\textbf{Implementation Details.}
Our method is implemented on two base models, FLUX.1-Kontext-dev and Qwen-Image, with 28 denoising steps for both. The db4 wavelet is used for the DWT, and high-frequency injection is applied during the first 30\% of the denoising steps.
For the adaptive injection strategy, FLUX.1 Kontext uses $\alpha_0=1.6$ and $\gamma=2.0$, while Qwen-Image uses $\alpha_0=2.0$ and $\gamma=1.6$. The path compensation mechanism operates with a period of $n=4$ steps for both models. For FLUX.1 Kontext, we additionally activate quality guidance during the final 30\% of denoising steps with $\lambda=0.3$.

\noindent\textbf{Baselines.}
We compare our method against seven representative baselines: FLUX.1-Kontext-dev~\cite{labs2025flux1kontextflowmatching}, Qwen-Image~\cite{wu2025qwenimagetechnicalreport}, Seedream 4.0~\cite{seedream2025seedream40nextgenerationmultimodal}, Nano Banana~\cite{comanici2025gemini25pushingfrontier}, MTC~\cite{MTC}, VINCIE~\cite{qu2025vincie}, and Bagel~\cite{deng2025bagel}. For open-source models, we use their official implementations with default settings. For closed-source models, we access them through their official APIs.

\noindent\textbf{Dataset.}
For quantitative evaluation, we collect 70 source images, equally divided between real-world photographs and high-quality synthetic images generated by FLUX.1-dev~\cite{flux2024}. For each source image, we employ Gemini 2.5 Pro~\cite{comanici2025gemini25pushingfrontier} to automatically generate a sequence of 10 progressive editing instructions, designed to simulate realistic multi-turn user interactions. These editing operations encompass five diverse categories: object manipulation, attribute modification, background replacement, style transfer, and action variation.

\noindent\textbf{Metrics.}
We employ a comprehensive evaluation framework combining traditional metrics with advanced VLM-based assessments. LPIPS~\cite{lpips2018} and CLIP-I~\cite{clip2021} are adopted to measure perceptual similarity and semantic alignment between edited and source images, respectively. Inspired by EdiVal-Agent~\cite{EdiVal}, we introduce three composite metrics powered by vision-language models: 
1) \textit{Instruction Following}: we leverage GPT-4o~\cite{openai2024gpt4ocard} to evaluate whether each edited image accurately fulfills its corresponding instruction; 
2) \textit{Consistency}: we employ DINOv2~\cite{oquab2023dinov2} and L1 distance to evaluate subject consistency, and use GPT-4o to assess background consistency; 
3) \textit{Quality}: we employ GPT-4o and HPSv3~\cite{hpsv3} to evaluate overall visual quality. Further details of the metric computation are provided in Appendix~\ref{sec:appendix_implementation}.

\subsection{Results}
\label{sec:results}

\noindent\textbf{Qualitative Evaluation.}
Figure~\ref{fig:qualitative} presents a qualitative comparison across 10 consecutive editing turns. As shown, Qwen-Image and FLUX.1 Kontext exhibit severe human body deformation and texture collapse as editing iterations accumulate. VINCIE, Seedream 4.0, and Bagel all fail to preserve fine-grained facial features, resulting in poor image quality with significant artifacts. While MTC successfully maintains overall image quality throughout the editing sequence, it demonstrates limited instruction-following capability, particularly for complex modifications involving background changes or human action alterations. Nano Banana exhibits strong subject consistency and editing fidelity; however, it introduces undesirable global color shifts across editing rounds, which is also reflected in its LPIPS scores in the quantitative evaluation. Our method substantially enhances the multi-turn editing capabilities of both Qwen-Image and FLUX.1 Kontext, achieving the best balance among instruction following, subject consistency, and overall perceptual quality. Notably, while Nano Banana represents the current state-of-the-art foundation model and substantially outperforms the open-source models (Qwen-Image and FLUX.1 Kontext), our method achieves comparable performance to Nano Banana in challenging multi-turn editing scenarios. Additional qualitative results can be found in Appendix~\ref{sec:appendix_qualitative}.

\input{tab_tex/quantitative}

\noindent\textbf{Quantitative Evaluation.}
Table~\ref{tab:quantitative} presents the quantitative results averaged over 10 consecutive editing turns. Seedream 4.0, VINCIE, and Bagel demonstrate substantially inferior performance in both perceptual quality and consistency metrics. While MTC achieves the highest perceptual quality scores, it exhibits the lowest instruction-following capability, indicating limited ability to execute complex editing instructions. As a state-of-the-art foundation model, Nano Banana consistently achieves high scores across all metrics, demonstrating strong balance between editing accuracy and content preservation.
Our method substantially enhances the base Qwen-Image and FLUX.1 Kontext models, particularly in image quality and content consistency. Notably, Qwen-Image with FreqEdit achieves superior performance across all three consistency metrics while maintaining competitive instruction-following capability. Compared to the base models, FreqEdit introduces only a marginal decline in instruction following (e.g., from 0.803 to 0.790 for FLUX.1 Kontext), demonstrating the effectiveness of our adaptive injection strategy and path compensation mechanism. Critically, this trade-off is well justified: the base models exhibit severe deformations and texture collapse (Figure~\ref{fig:qualitative}), rendering their outputs practically unusable, whereas FreqEdit substantially improves visual fidelity while retaining strong instruction-following capability. Additional quantitative results are provided in Appendix~\ref{sec:appendix_quantitative}.

\input{fig_tex/ablation1}

\noindent\textbf{Human Preference Evaluation.}
We conduct a human preference study where participants rank methods based on multi-turn editing results. As shown in the last column of Table~\ref{tab:quantitative}, our Qwen-Image + FreqEdit achieves the highest preference score, followed by Nano Banana and FLUX.1 Kontext + FreqEdit. Notably, FreqEdit-enhanced variants consistently outperform their native counterparts, confirming that our approach improves user-perceived quality. These rankings align well with our qualitative and quantitative findings, validating the effectiveness and consistency of our approach. Additional details of the user study are provided in Appendix~\ref{sec:appendix_user_study}.

\subsection{Ablation Study}
To validate the effectiveness of each component in our framework, we conduct ablation studies by removing each individual component. Figure~\ref{fig:ablation} presents a visual comparison of the variant models. Without the adaptive injection strategy, the model fails to properly execute complex semantic edits, such as background transformations. Removing the path compensation mechanism sometimes introduces visible ghosting artifacts caused by conflicting optimization signals between the editing and reference velocity fields. Without quality guidance, FLUX.1 Kontext exhibits severe noise artifacts across multiple editing iterations. These results demonstrate that all three components are essential for achieving robust multi-turn editing.

\noindent\textbf{Analysis of Frequency Component Selection.}
To validate our design choice of injecting only high-frequency (HF) features from the reference image, we compare against two alternatives: injecting only low-frequency (LF) features and injecting all frequency components (HF + LF). As shown in Figure~\ref{fig:ablation}(d), injecting all components frequently leads to edit failure, as the substantially greater semantic energy carried by LF components relative to HF components conflicts with the editing instructions. Injecting only LF features, on the other hand, fails to prevent subject deformation and similarly results in edit failure, confirming that HF components are the key to preventing subject deformation during multi-turn editing.

%% file: fig_tex/qualitative.tex
\begin{figure*}[htb]
    \centering
    \vspace{-8pt}
    \includegraphics[width=\textwidth]{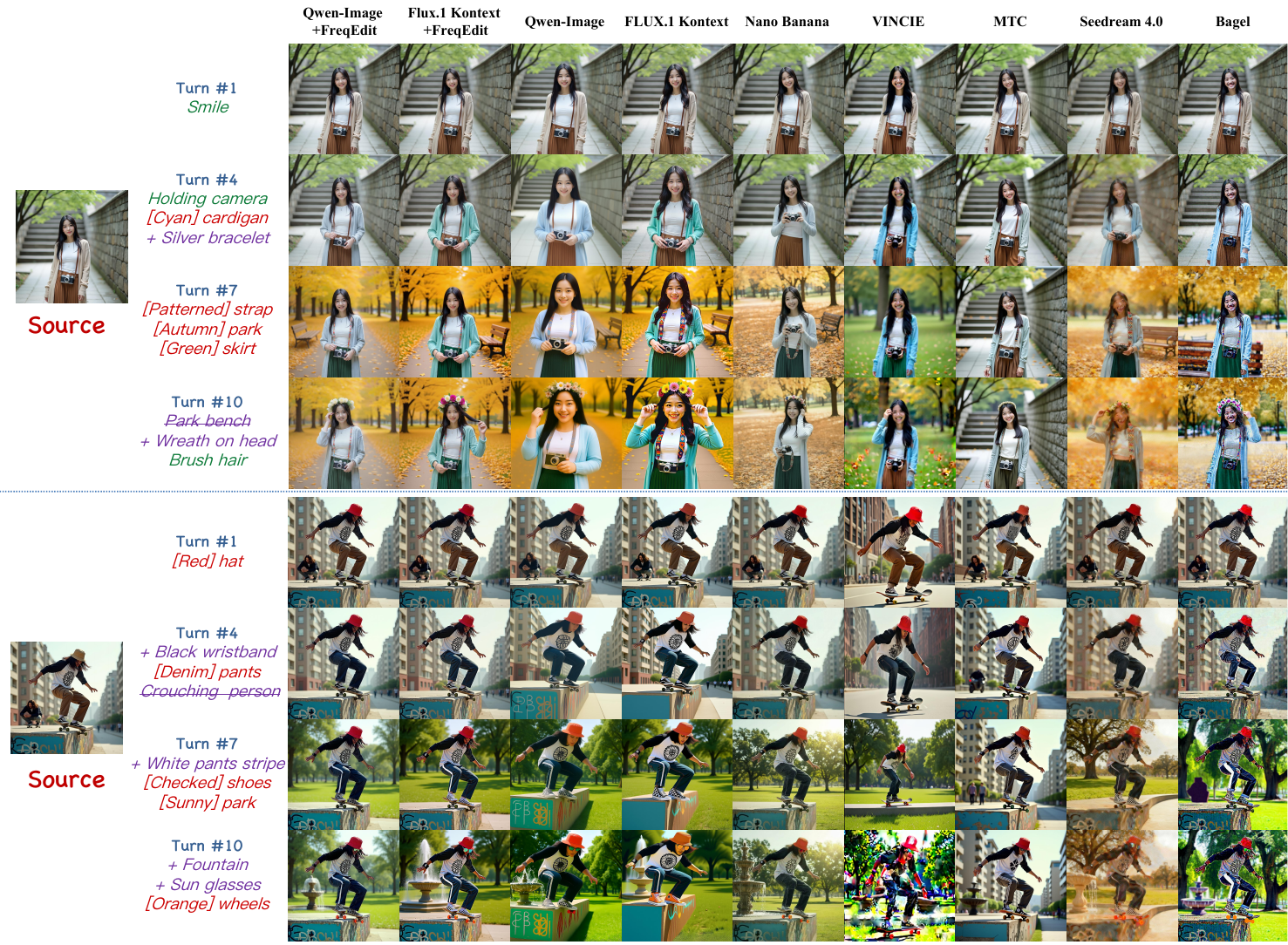}
\caption{\textbf{Qualitative comparison of iterative editing.} Compared to FLUX.1 Kontext~\cite{labs2025flux1kontextflowmatching}, Qwen-Image~\cite{wu2025qwenimagetechnicalreport}, Seedream 4.0~\cite{seedream2025seedream40nextgenerationmultimodal}, Nano Banana~\cite{comanici2025gemini25pushingfrontier}, MTC~\cite{MTC}, VINCIE~\cite{qu2025vincie}, and Bagel~\cite{deng2025bagel}, our method achieves a better balance among instruction following, subject consistency, and overall perceptual quality. Please zoom in for a better view.}
\vspace{-8pt} 
\label{fig:qualitative}
\end{figure*}

%% file: tab_tex/quantitative.tex
\begin{table*}[!t]
  \centering
  \small
  \setlength{\tabcolsep}{4pt}
\caption{\textbf{Quantitative results across 10 sequential edits.} Our method demonstrates stable performance across all metrics throughout the editing sequence. ``Instr.'' and ``Cons.'' denote instruction-following and consistency metrics, respectively. The best results are highlighted in bold, while the second-best results are underlined.}
    \vspace{-5pt}
  \resizebox{\textwidth}{!}{
  \begin{tabular}{
    l
    *{5}{c}
    *{5}{c}
    *{5}{c}
    *{6}{c}
  }
  \toprule
  \multirow{2}{*}{\textbf{Methods}} &
  \multicolumn{5}{c}{\textbf{Turn 1}} &
  \multicolumn{5}{c}{\textbf{Turn 4}} &
  \multicolumn{5}{c}{\textbf{Turn 7}} &
  \multicolumn{6}{c}{\textbf{Turn 10}} \\
  \cmidrule(lr){2-6}\cmidrule(lr){7-11}\cmidrule(lr){12-16}\cmidrule(lr){17-22}
  & \textbf{\makecell[c]{CLIP-I$\uparrow$}} 
  & \textbf{\makecell[c]{LPIPS$\downarrow$}} 
  & \textbf{\makecell[c]{Instr.$\uparrow$}} 
  & \textbf{\makecell[c]{Cons.$\uparrow$}} 
  & \textbf{\makecell[c]{Quality$\uparrow$}}
  & \textbf{\makecell[c]{CLIP-I$\uparrow$}} 
  & \textbf{\makecell[c]{LPIPS$\downarrow$}} 
  & \textbf{\makecell[c]{Instr.$\uparrow$}} 
  & \textbf{\makecell[c]{Cons.$\uparrow$}} 
  & \textbf{\makecell[c]{Quality$\uparrow$}}
  & \textbf{\makecell[c]{CLIP-I$\uparrow$}} 
  & \textbf{\makecell[c]{LPIPS$\downarrow$}} 
  & \textbf{\makecell[c]{Instr.$\uparrow$}} 
  & \textbf{\makecell[c]{Cons.$\uparrow$}} 
  & \textbf{\makecell[c]{Quality$\uparrow$}}
  & \textbf{\makecell[c]{CLIP-I$\uparrow$}} 
  & \textbf{\makecell[c]{LPIPS$\downarrow$}} 
  & \textbf{\makecell[c]{Instr.$\uparrow$}} 
  & \textbf{\makecell[c]{Cons.$\uparrow$}} 
  & \textbf{\makecell[c]{Quality$\uparrow$}}
  & \textbf{\makecell[c]{Human$\uparrow$}} \\
  \midrule
  Bagel & 0.953 & 0.135 & 0.799 & 0.894 & 0.728 & 0.896 & 0.321 & 0.770 & 0.804 & 0.630 & 0.857 & 0.456 & 0.768 & 0.750 & 0.553 & 0.822 & 0.546 & 0.768 & 0.709 & 0.494&4.830 \\
  MTC & 0.924 & 0.355 & 0.500 & 0.721 & \textbf{0.798} & 0.909 & 0.405 & 0.555 & 0.750 & \textbf{0.799} & 0.896 & 0.431 & 0.558 & 0.750 & \textbf{0.793} & 0.886 & 0.449 & 0.554 & 0.746 & \textbf{0.790}&6.246 \\
  Nano Banana & \underline{0.972} & 0.159 & 0.805 & 0.907 & 0.767 & \underline{0.944} & 0.295 & \underline{0.808} & 0.866 & 0.758 & \underline{0.919} & 0.396 & \underline{0.822} & \underline{0.832} & 0.743 & \underline{0.893} & 0.472 & \underline{0.835} & \underline{0.806} & \underline{0.731}&\underline{7.271} \\
  Seedream 4.0 & 0.967 & 0.180 & \textbf{0.865} & 0.881 & 0.732 & 0.914 & 0.394 & \textbf{0.832} & 0.813 & 0.646 & 0.861 & 0.541 & \textbf{0.827} & 0.759 & 0.575 & 0.820 & 0.635 & \textbf{0.849} & 0.720 & 0.527 &4.241\\
  VINCIE & 0.944 & 0.359 & 0.786 & 0.781 & 0.662 & 0.904 & 0.479 & 0.799 & 0.742 & 0.623 & 0.872 & 0.565 & 0.740 & 0.694 & 0.571 & 0.846 & 0.618 & 0.697 & 0.654 & 0.524 &4.148\\
\rowcolor{kontextblue}
  FLUX.1 Kontext + FreqEdit & \cellcolor{kontextblueDeep}\underline{0.972} & \cellcolor{kontextblueDeep}\underline{0.115} & 0.784 & \cellcolor{kontextblueDeep}\underline{0.921} & 0.758 & \cellcolor{kontextblueDeep}0.941 & \cellcolor{kontextblueDeep}\underline{0.218} & 0.776 & \cellcolor{kontextblueDeep}\underline{0.872} & \cellcolor{kontextblueDeep}0.747 & \cellcolor{kontextblueDeep}0.912 & \cellcolor{kontextblueDeep}\underline{0.330} & 0.771 & \cellcolor{kontextblueDeep}0.831 & \cellcolor{kontextblueDeep}0.726 & \cellcolor{kontextblueDeep}0.884 & \cellcolor{kontextblueDeep}\underline{0.418} & 0.790 & \cellcolor{kontextblueDeep}0.798 & \cellcolor{kontextblueDeep}0.712&\cellcolor{kontextblueDeep}6.910 \\
  
\rowcolor{kontextblue}
  FLUX.1 Kontext & 0.966 & 0.222 & \cellcolor{kontextblueDeep}0.798 & 0.902 & \cellcolor{kontextblueDeep}0.764 & 0.927 & 0.365 &\cellcolor{kontextblueDeep} 0.791 & 0.843 & 0.739 & 0.889 & 0.468 & \cellcolor{kontextblueDeep}0.791 & 0.799 & 0.706 & 0.854 & 0.542 & \cellcolor{kontextblueDeep}0.803 & 0.762 & 0.681& 4.920\\
\rowcolor{qwenred}
  Qwen-Image + FreqEdit & \cellcolor{qwenredDeep}\textbf{0.973} & \cellcolor{qwenredDeep}\textbf{0.097} & 0.790 & \cellcolor{qwenredDeep}\textbf{0.924} & 0.768 & \cellcolor{qwenredDeep}\textbf{0.948} & \cellcolor{qwenredDeep}\textbf{0.192} & 0.774 & \cellcolor{qwenredDeep}\textbf{0.877} & \cellcolor{qwenredDeep}\underline{0.761} & \cellcolor{qwenredDeep}\textbf{0.923} & \cellcolor{qwenredDeep}\textbf{0.291} & 0.768 & \cellcolor{qwenredDeep}\textbf{0.840} & \cellcolor{qwenredDeep}\underline{0.745} & \cellcolor{qwenredDeep}\textbf{0.897} & \cellcolor{qwenredDeep}\textbf{0.374} & 0.784 & \cellcolor{qwenredDeep}\textbf{0.807} & \cellcolor{qwenredDeep}0.729 &\cellcolor{qwenredDeep}\textbf{7.393}\\
\rowcolor{qwenred}
  Qwen-Image & 0.969 & 0.236 & \cellcolor{qwenredDeep}\underline{0.806} & 0.902 & \cellcolor{qwenredDeep}\underline{0.772} 
  
  & 0.931 & 0.393 & \cellcolor{qwenredDeep}0.785 & 0.841 & 0.752 & 0.898 & 0.494 & \cellcolor{qwenredDeep}0.795 & 0.800 & 0.731 & 0.871 & 0.566 & \cellcolor{qwenredDeep}0.809 & 0.767 & 0.713&5.177 \\
  \bottomrule
  \end{tabular}
  }
  \label{tab:quantitative}
  \vspace{-10pt}
\end{table*}

%% file: fig_tex/ablation1.tex
\begin{figure}[tb]
    \centering
    \includegraphics[width=\linewidth]{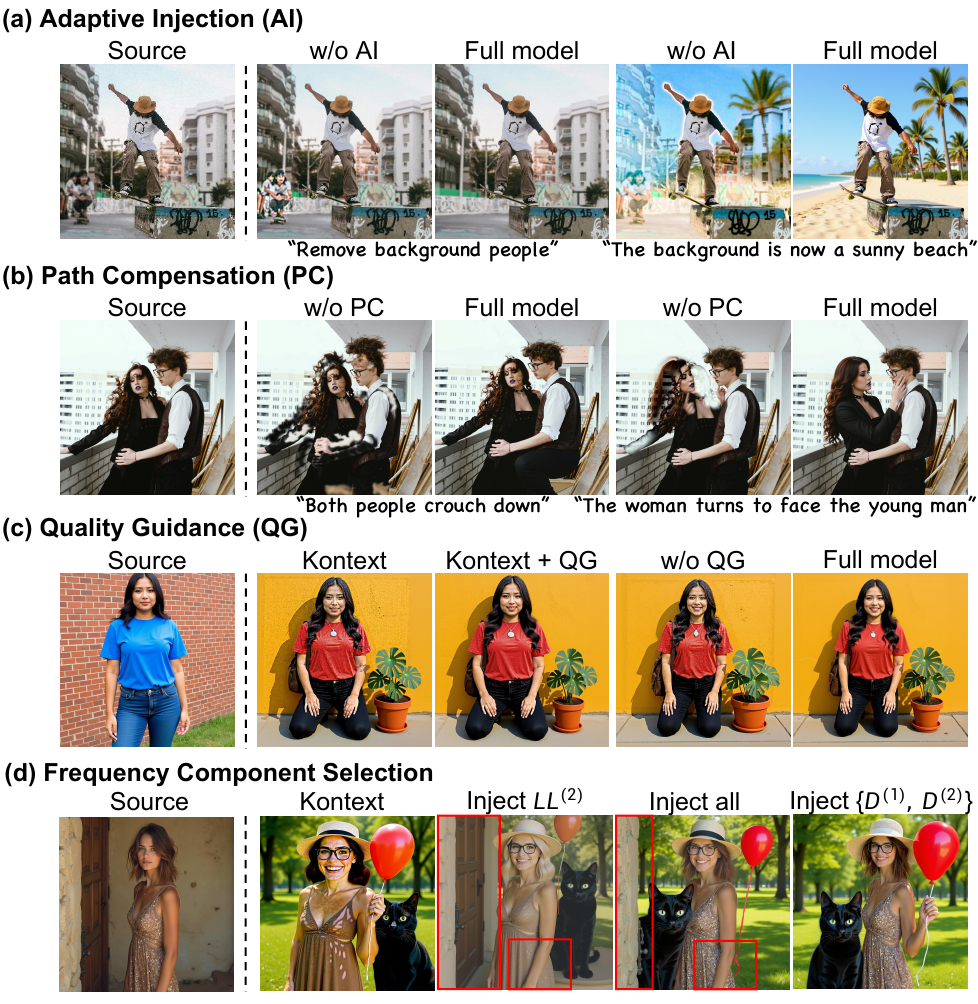}
    \caption{\textbf{Ablation study.} (a) Without the adaptive injection strategy, the model fails to perform background transformation and subject removal. (b) Removing the path compensation mechanism introduces visible ghosting artifacts. (c) Without quality guidance, FLUX.1 Kontext exhibits severe noise artifacts after several editing iterations. (d) Injecting all features or low-frequency components, rather than selectively fusing high-frequency features, leads to semantic leakage and editing failures.}
    \vspace{-25pt}
    \label{fig:ablation}
\end{figure}

%% file: 6_conclusion.tex
\section{Conclusions and Limitations}
In this work, we address the critical challenge of maintaining visual consistency in multi-turn image editing. Through systematic analysis, we identify that the progressive degradation of high-frequency features is the fundamental cause of quality deterioration across editing turns. We introduce FreqEdit, a training-free framework that mitigates this degradation by selectively injecting reference velocity components. While FreqEdit effectively preserves high-frequency fidelity throughout multi-turn editing, it inherently depends on the high-frequency content present in the source image, and may exhibit reduced effectiveness when a single edit spans a large spatial region. Additional limitations and future work are discussed in Appendix~\ref{sec:discussion_limitation}.

\clearpage

\section*{Acknowledgments}
This work is supported by National Natural Science Foundation of China (No.~62302535, No.~61902087, and No.~62176223) and Zhuhai Basic and Applied Basic Research Foundation (No.~2320004002745).

%% file: 7_appendix.tex
\clearpage
\appendix

\twocolumn
[
\begin{center}
{\Large\textbf{Appendices}} 
\end{center}
]
\vspace{20pt}

\section*{Contents}

\apptocsection{A}{Reference Velocity Field Formulation}{sec:appendix_derivation}
\vspace{7pt}

\apptocsection{B}{VLM-Based Evaluation Metric Details}{sec:appendix_implementation}
\vspace{7pt}

\apptocsection{C}{Additional Qualitative Results}{sec:appendix_qualitative}
\vspace{7pt}

\apptocsection{D}{Additional Quantitative Results}{sec:appendix_quantitative}
\vspace{7pt}

\apptocsection{E}{User Study Details}{sec:appendix_user_study}
\vspace{7pt}

\apptocsection{F}{Additional Ablation Study}{sec:appendix_ablation}
\vspace{7pt}

\apptocsection{G}{Latent-Space Frequency Alignment}{sec:appendix_alignment}
\vspace{7pt}

\apptocsection{H}{Path Compensation Visualization}{sec:visualization_PC}
\vspace{7pt}

\apptocsection{I}{Multi-Turn Editing vs. Combined Prompts}{sec:combined_vs_multiturn}
\vspace{7pt}

\apptocsection{J}{More Discussion on Limitations and Future Work}{sec:discussion_limitation}
\vspace{7pt}

\apptocsection{K}{Multi-Turn Editing Instruction Generation}{sec:appendix_templates}

\vspace{1em}

\section{Reference Velocity Field Formulation}
\label{sec:appendix_derivation}
We derive the reference velocity field in Eq.~(\ref{eq:v_ref}) from the Euler discretization of the rectified flow ODE. The update rule of Euler discretization is:
\begin{equation}
Z_{t_{i-1}} = Z_{t_i} + (t_{i-1} - t_i)\, v_\theta(Z_{t_i}, t_i, \mathbf{c}),
\end{equation}
which provides a finite-difference approximation of the instantaneous velocity:
\begin{equation}
v_\theta(Z_{t_i}, t_i, \mathbf{c}) = 
\frac{Z_{t_{i-1}} - Z_{t_i}}{t_{i-1} - t_i}.
\label{eq:appendix_single_step_no_section}
\end{equation}
For brevity, we denote $v_{t_j} \equiv v_\theta(Z_{t_j}, t_j, \mathbf{c})$. Applying the update rule iteratively from step $i$ down to step $1$ yields:
\begin{align}
Z_{t_{i-1}} - Z_{t_i} &= (t_{i-1} - t_i)\, v_{t_i}, \\
Z_{t_{i-2}} - Z_{t_{i-1}} &= (t_{i-2} - t_{i-1})\, v_{t_{i-1}}, \\
&\;\;\vdots \nonumber \\
Z_{t_0} - Z_{t_1} &= (t_0 - t_1)\, v_{t_1}.
\end{align}
Summing over all remaining steps from $i$ down to $1$ produces the telescoping relation:
\begin{equation}
Z_{t_0} - Z_{t_i} 
= \sum_{j=1}^{i} (t_{j-1} - t_j)\, v_{t_j}.
\label{eq:appendix_telescoping_no_section}
\end{equation}
Since the terminal state $Z_{t_0}$ corresponds to the context image $Z^{\text{ref}}_0$, we have:
\begin{equation}
Z^{\text{ref}}_0 - Z_{t_i}
= \sum_{j=1}^{i} (t_{j-1} - t_j)\, v_{t_j}.
\end{equation}
We approximate the velocity field to be constant over the remaining interval [$t_0$, $t_i$], setting it equal to $v^{\text{ref}}_{ti}$. Under this constant-velocity approximation:
\begin{equation}
Z^{\text{ref}}_0 - Z_{t_i}
= \sum_{j=1}^{i} (t_{j-1} - t_j)\, v^{\text{ref}}_{t_i}
= (t_0 - t_i)\, v^{\text{ref}}_{t_i},
\end{equation}
which directly yields:
\begin{equation}
v^{\text{ref}}_{t_i} = \frac{Z^{\text{ref}}_0 - Z_{t_i}}{t_0 - t_i},
\label{eq:appendix_v_ref_final}
\end{equation}
which recovers Eq.~(\ref{eq:v_ref}) from the main text.
This expression represents the average velocity required to traverse from $Z_{t_i}$ to the reference image $Z^{\text{ref}}_0$ over the remaining time $(t_0 - t_i)$. This formulation is analogous to the average velocity approach proposed in MeanFlow~\cite{meanflow}. Geometrically, this induces a straight-line trajectory in the latent space that preserves the high-frequency features of the context image.

\section{VLM-Based Evaluation Metric Details}
\label{sec:appendix_implementation}
This section details our VLM-based evaluation metrics, including instruction following, visual consistency, and perceptual quality.

\vspace{5pt}
\noindent\textbf{Instruction Following.}
For each editing turn $k$, we evaluate how well the generated image $X^{[k+1]}$ follows the corresponding instruction $p^{[k]}$ by comparing it against the previous image $X^{[k]}$. We employ GPT-4o as the evaluator with a carefully designed prompt that instructs the model to focus exclusively on instruction adherence. The evaluation prompt consists of three components: (1) a message emphasizing objective evaluation, (2) the editing instruction $p^{[k]}$, and (3) the source image $X^{[k]}$ and the edited image $X^{[k+1]}$. The model returns a score in $[0, 1]$, where 1 indicates perfect instruction fulfillment and 0 indicates complete failure. 
The evaluation prompt is specifically designed to assess instruction adherence in isolation, explicitly instructing the evaluator to disregard visual quality.

\vspace{5pt}
\noindent\textbf{Visual Consistency.}
Our consistency metric comprises subject consistency and background consistency. For subject consistency, instead of detecting all objects indiscriminately, we first leverage GPT-4o to infer which subjects should be present after each editing operation. Specifically, we provide GPT-4o with: (1) the original image $X^{[1]}$, and (2) the cumulative editing instructions $\{p^{[1]}, p^{[2]}, \ldots, p^{[k]}\}$. The model reasons about the editing semantics and outputs a list of expected subject categories $\mathcal{S}^{[k+1]} = \{s_1, s_2, \ldots, s_n\}$ (e.g., ``person'', ``blue car'', ``tree''). This inference-based approach enables editing-aware subject tracking, as opposed to naive object detection.

For each inferred subject category $s_i \in \mathcal{S}^{[k+1]}$, we use it as the text prompt for GroundingDINO~\cite{liu2023grounding} to localize the corresponding object instance in images $X^{[1]}$, $X^{[k]}$, and $X^{[k+1]}$. We set the box confidence threshold to 0.25, the text matching threshold to 0.25, and enable \texttt{keep\_top1\_per\_label} to retain only the highest-confidence detection per subject category, thereby reducing false positives. We denote the set of successfully detected object instances as $\mathcal{O}$.

We compute consistency for object instances that are successfully detected in both source and target images. Specifically, we measure: (1) original-to-current consistency: consistency of objects present in both $X^{[1]}$ and $X^{[k+1]}$, and (2) previous-to-current consistency: consistency of objects present in both $X^{[k]}$ and $X^{[k+1]}$.

For each common object instance $o \in \mathcal{O}$, we extract its region-of-interest (ROI) based on the detected bounding box and compute two types of features: (1) DINOv2 features: We resize each ROI to $224 \times 224$ pixels and feed it to DINOv2~\cite{oquab2023dinov2}. Consistency is measured via cosine similarity:
\begin{equation}
    \text{sim}_{\text{DINOv2}}(o) = \frac{\mathbf{f}_{\text{src}}(o) \cdot \mathbf{f}_{\text{tgt}}(o)}{\|\mathbf{f}_{\text{src}}(o)\| \, \|\mathbf{f}_{\text{tgt}}(o)\|},
\end{equation}
where $\mathbf{f}_{\text{src}}(o)$ and $\mathbf{f}_{\text{tgt}}(o)$ are DINOv2 features of object $o$ in the source and target images, respectively. (2) L1 pixel distance: We compute the normalized L1 distance between the resized ROI patches (with pixel values normalized to $[0,1]$) and convert it to a similarity score:
\begin{equation}
    \text{sim}_{\text{L1}}(o) = 1 - \frac{1}{HWC} \sum_{h,w,c} | P_{\text{src}}^{(o)}(h,w,c) - P_{\text{tgt}}^{(o)}(h,w,c) |,
\end{equation}
where $P_{\text{src}}^{(o)}$ and $P_{\text{tgt}}^{(o)}$ are the RGB patches of object $o$ with spatial dimensions $H \times W$ and $C=3$ color channels.

For each similarity type, the subject consistency score is computed by averaging over all common object instances:
\begin{equation}
    \text{Consistency}_{\text{subject}}^{\text{type}} = \frac{1}{|\mathcal{O}|} \sum_{o \in \mathcal{O}} \text{sim}_{\text{type}}(o),
\end{equation}
where $\mathcal{O}$ denotes the set of common object instances detected in both source and target images, and $\text{type} \in \{\text{DINOv2}, \text{L1}\}$.

For background consistency, when the editing instruction does not involve background modification (as determined by intent parsing), we evaluate background preservation using GPT-4o. The evaluation protocol consists of two stages: (1) we explicitly instruct the model to focus only on background regions and ignore foreground objects identified during subject consistency evaluation, and (2) the model assesses background consistency across four dimensions, including \textit{layout} (preservation of spatial arrangement), \textit{texture} (consistency of material properties), \textit{lighting} (color temperature and shadow consistency), and \textit{artifacts} (absence of boundary artifacts or distortions).

\vspace{5pt}
\noindent\textbf{Perceptual Quality.}
We assess the visual quality of each generated image $X^{[k+1]}$ using two complementary quality models. We obtain GPT-4o-based scores by prompting the model to evaluate six dimensions, including \textit{aesthetics}, \textit{realism}, \textit{sharpness}, \textit{exposure}, \textit{artifacts}, and \textit{composition}, each scored in $[0,1]$, along with an overall quality score. We then compute HPSv3~\cite{hpsv3} scores as a learned perceptual quality metric.

\section{Additional Qualitative Comparisons}
\label{sec:appendix_qualitative}
Figures~\ref{fig:qual_exp_1} to \ref{fig:qual_exp_6} present additional qualitative comparisons across diverse multi-turn editing scenarios, including complex background modifications, fine-grained attribute changes, action and pose alterations, object manipulations and style transfers. These extended results consistently reveal the limitations of existing approaches. Specifically, Qwen-Image and FLUX.1 Kontext exhibit progressive quality degradation with accumulated artifacts and body deformations. VINCIE, Seedream 4.0, and Bagel fail to preserve facial details and introduce significant visual artifacts. While MTC maintains reasonable image quality, it demonstrates limited instruction-following capability for complex edits. Nano Banana achieves competitive performance but suffers from noticeable color shifts across editing iterations. Across all scenarios, our method achieves the superior overall performance, successfully balancing accurate instruction following, robust subject consistency, and high perceptual quality.

\section{Additional Quantitative Results}
\label{sec:appendix_quantitative}

\input{tab_tex/additional_quantitative_average}

\input{fig_tex/additional_quantitative_fig}

To provide a more comprehensive evaluation, we present results on three additional metrics: PSNR, SSIM~\cite{wang2004ssim}, and DINO-Sim~\cite{oquab2023dinov2}. PSNR measures pixel-level reconstruction quality between the edited and original images. SSIM evaluates structural information preservation, including luminance, contrast, and texture patterns. DINO-Sim measures high-level semantic consistency via cosine similarity between DINO features. All metrics are computed between the edited image $X^{[k+1]}$ at turn $k$ and the original unedited image $X^{[1]}$. 

Table~\ref{tab:additional_quantitative} presents the cumulative mean of PSNR, SSIM, and DINO-Sim across progressive editing turns for all compared methods. FreqEdit demonstrates substantial and consistent improvements when integrated with different base models. Both FLUX.1 Kontext and Qwen-Image exhibit significant gains across all three metrics when equipped with FreqEdit. As discussed in Section~\ref{sec:results}, MTC achieves high PSNR values due to its limited ability to execute editing instructions, resulting in edited images that remain largely unchanged from the original. Notably, Qwen-Image + FreqEdit achieves the best results across most metrics and editing turns.

We further provide quantitative results at each individual editing turn relative to the original image in Figure~\ref{fig:additional_quantitative_fig}. As shown, FreqEdit-equipped models consistently maintain higher similarity scores at every turn, with improvements ranging from 1--3~dB in PSNR, 0.05--0.15 in SSIM, and 0.05--0.1 in DINO-Sim across different turns. Importantly, FreqEdit-equipped models not only achieve higher absolute scores but also exhibit more stable degradation patterns throughout the editing sequence, demonstrating superior consistency preservation during iterative editing.

\section{User Study Details}
\label{sec:appendix_user_study}

We conduct a comprehensive user study to evaluate the perceptual quality of our method against seven baseline approaches. This section provides detailed information on our study protocol and scoring mechanism. 

We design an online questionnaire-based user study where participants rank multiple methods based on their editing results. Participants are instructed to comprehensively consider three key aspects when making their judgments: \textit{aesthetics} (visual appeal), \textit{instruction following} (accuracy in executing given instructions), and \textit{consistency} (preservation of unedited regions and object identity across turns). For each question in the survey, participants are shown: 1) the source (unedited) image $X^{[1]}$, 2) edited results at three intermediate turns: $X^{[5]}$, $X^{[8]}$, and $X^{[11]}$, and 3) all editing instructions $\{p^{[1]}, p^{[2]}, \ldots, p^{[10]}\}$ applied sequentially. We collect a total of 60 completed survey responses.

Given the cognitive difficulty of ranking all 9 methods simultaneously, we adopt a random sampling strategy. For each question, we randomly select 7 out of 9 methods and present their results to participants, who then rank these methods from best (rank 1) to worst (rank 7) based on overall quality. This choice of 7 methods balances cognitive load with questionnaire efficiency and reliability.

To obtain a unified score for each method, we convert the collected rankings to a 9-point scale. Let $n_k$ denote the number of times a method receives rank $k$ (where $k \in \{1, 2, \ldots, 7\}$). We assign descending weights to each rank using the weight vector:
\begin{equation}
\mathbf{w} = [9, 8, 7, 6, 5, 4, 3],
\end{equation}
where rank 1 receives weight 9 and rank 7 receives weight 3. The final score $S$ for a given method is computed as:
\begin{equation}
S = \frac{9n_1 + 8n_2 + 7n_3 + 6n_4 + 5n_5 + 4n_6 + 3n_7}{\sum_{k=1}^{7} n_k}.
\label{eq:user_study_score}
\end{equation}
This formulation ensures that methods consistently ranked higher receive proportionally higher scores, with the weight assignment reflecting relative preference while mapping rankings to an intuitive 9-point scale.

\section{Additional Ablation Study}
\label{sec:appendix_ablation}

\input{fig_tex/additional_ablation_AI}

\input{fig_tex/additional_ablation_PC}

\input{fig_tex/additional_ablation_QG}

\input{fig_tex/additional_ablation_hyperparameters}

To validate the effectiveness of each component in our framework, we conduct ablation studies by systematically removing individual components. We evaluate three variants: without adaptive injection strategy, without path compensation mechanism, and without quality guidance. In addition, we analyze the sensitivity of FreqEdit to its three key hyperparameters ($\alpha_0$, $\gamma$, and $\lambda$), varying each individually while fixing the others to their default values.

\vspace{5pt}
\noindent\textbf{Adaptive Injection.} 
Figure~\ref{fig:additional_ablation_AI} presents qualitative comparisons validating our adaptive injection strategy (Section~\ref{sec:adaptive}). 
Without adaptive injection, uniform injection strength often leads to over-preservation artifacts in semantically modified regions. 
In the top row, the model overly preserves the original background and gate geometry, preventing full transition to a sunny beach scene (column 2) and inhibiting the reconstruction of a white wooden picket gate (column 4). Similarly, in the bottom row, large portions of the original background remain intact instead of forming rolling green hills (column 2), and the horse's coat retains much of its initial appearance, failing to exhibit a coherent dapple-gray pattern (column 4). In contrast, our adaptive injection approach enables faithful editing in modified regions while preserving details elsewhere.

\vspace{5pt}
\noindent\textbf{Path Compensation.}
Figure~\ref{fig:additional_ablation_PC} demonstrates the effectiveness of our path compensation strategy (Section~\ref{sec:path_compensation}). Without path compensation, high injection strength leads to ghosting artifacts where visual elements from both editing and reference velocity fields appear simultaneously. The kickflip scene (top row, second column) shows duplicated skateboarder poses with overlapping limbs, the mid-air jump (fourth column) produces ghost figures, the cap toss (bottom row, second column) exhibits duplicated body configurations, and the parent addition (fourth column) displays semi-transparent overlapping figures. 
These results confirm that path compensation is crucial for stabilizing the editing trajectory and preventing ghosting artifacts.

\vspace{5pt}
\noindent\textbf{Quality Guidance.}
The quality guidance mechanism (Section~\ref{sec:quality_preservation}) addresses noise accumulation across editing iterations, particularly for models like FLUX.1 Kontext. Figure~\ref{fig:additional_ablation_QG} demonstrates its effectiveness through two representative examples after multiple editing turns.
We compare four configurations: native model baseline, native model with quality guidance, our framework without quality guidance (w/o QG), and our complete model. The native FLUX.1 Kontext (column 2) exhibits pronounced noise artifacts, particularly in skin textures. Applying quality guidance alone (column 3) substantially reduces these artifacts by blending the editing velocity with an auxiliary velocity from the original high-quality image during final denoising steps. When integrated into our complete framework, quality guidance (column 5) further refines visual quality compared to the variant without it (column 4), producing clean results that maintain both high-frequency details and low noise levels. 

\vspace{5pt}
\noindent\textbf{Hyperparameter Sensitivity.}
We analyze the sensitivity of FreqEdit to its three key hyperparameters on FLUX.1 Kontext, varying each individually while keeping the others fixed at their default values ($\alpha_0 = 1.6$, $\gamma = 2.0$, $\lambda = 0.3$). As shown in Figure~\ref{fig:ablation_hyperparameters}, $\alpha_0$ and $\gamma$ jointly govern the magnitude and spatial distribution of high-frequency feature injection, while $\lambda$ independently controls the quality guidance strength applied during the final denoising steps.

$\alpha_0$ controls the overall injection magnitude in Eq.~\eqref{eq:adaptive_injection}. When $\alpha_0$ is too small, the injected high-frequency signal is insufficient to counteract progressive degradation, leading to subject deformation in later editing turns. Conversely, when $\alpha_0$ is too large, the reference velocity dominates the editing velocity, causing editing failure where the intended semantic changes are suppressed. Our method is robust for $\alpha_0 \in [1.5, 2.5]$.

$\gamma$ jointly controls the overall injection strength and the contrast of the spatially-adaptive injection map (Eq.~\eqref{eq:adaptive_injection}). A small $\gamma$ yields both weaker injection and a low-contrast map that fails to sufficiently differentiate between semantically stable and modified regions, resulting in subject deformation due to inadequate high-frequency reinforcement. A large $\gamma$ amplifies injection strength and sharpens spatial boundaries, which can suppress desired semantic transformations and introduce visible transition artifacts. Our method is robust for $\gamma \in [1.5, 2.5]$.

$\lambda$ controls the blending strength of the quality guidance mechanism (Eq.~\eqref{eq:quality_guidance}), which is applied during the final denoising steps to suppress accumulated noise artifacts. When $\lambda$ is too small, residual noise artifacts remain visible in the output, particularly in fine-grained regions such as skin textures. When $\lambda$ is too large, the auxiliary velocity exerts excessive influence, causing over-saturation in the output image. Our method is robust for $\lambda \in [0.2, 0.4]$.

\input{fig_tex/frequency_alignment}

\section{Latent-Space Frequency Alignment}
\label{sec:appendix_alignment}

 FreqEdit operates on high-frequency components extracted from the latent velocity field via DWT. A key assumption underlying this design is that frequency decomposition in the latent space exhibits consistent behavior with that in the image space. To empirically validate this assumption, we apply 2-level DWT to the latent representation of a source image and selectively reconstruct images from individual frequency bands via VAE decoding.  
 
 As shown in Figure~\ref{fig:frequency_alignment}, the reconstructed images reveal a clear correspondence between latent-space frequency bands and their visual counterparts in image space. The second-level low-frequency component $LL^{(2)}$ captures the global structure and coarse appearance of the scene, including the overall layout, color distribution, and subject silhouette, albeit with noticeable blurring. The second-level high-frequency component $D^{(2)}$ encodes coarser-scale structural edges and textural patterns, such as body contours and clothing boundaries, without retaining global color information. Combining $LL^{(2)}$ and $D^{(2)}$ recovers a substantially sharper reconstruction that approximates the full image, confirming that these two levels together account for the dominant visual content. The first-level high-frequency component $D^{(1)}$ captures the finest-grained details, including hair strands, fabric textures, and subtle edge structures. These observations confirm that latent-space frequency components behave consistently with their image-space counterparts, providing a principled justification for performing high-frequency feature injection directly in the latent space.

\input{fig_tex/visualization_PC}

\section{Path Compensation Visualization}
\label{sec:visualization_PC}
While Section~\ref{sec:path_compensation} establishes the mathematical equivalence between the path compensation trajectory and the editing velocity, we provide here a step-level visualization to intuitively demonstrate how path compensation resolves the semantic conflict introduced by high-frequency injection during denoising.

Figure~\ref{fig:visualization_PC} visualizes the latent state at an intermediate denoising timestep $t_i$. Before applying path compensation, the corrected latent $Z_{t_i}$ exhibits residual structure from the high-frequency injection that conflicts with the editing objective, as highlighted by the red box. The trajectory buffer ${B}$ (Eq.~\eqref{eq:trajectory_buffer}) accumulates the per-step velocity divergence weighted by the timestep interval:
\begin{equation*}
    {B} \leftarrow {B} + (t_{i-1} - t_i) \cdot \Delta v_{t_i},
    \quad \text{where} \quad \Delta v_{t_i} = v^{\text{edit}}_{t_i} - v^{\text{corr}}_{t_i}.
\end{equation*}
As shown in the heatmap, ${B}$ carries a structured correction signal in the region (blue box) corresponding to the residual structure highlighted in the red box, counteracting the conflict introduced by high-frequency injection. Adding ${B}$ to the latent effectively suppresses this residual structure, yielding a clean latent aligned with the intended edit. This visualization confirms that path compensation periodically realigns the denoising trajectory with the editing objective while preserving the high-frequency information injected in preceding steps.

\input{fig_tex/combined_vs_multiturn}

\section{Multi-Turn Editing vs. Combined Prompts}
\label{sec:combined_vs_multiturn}

A natural question arises as to whether multi-turn editing can be replaced by combining all instructions into a single prompt, which would eliminate the need for iterative processing. While combining instructions may work in simple cases, we argue that multi-turn editing is both practically necessary and technically preferable for the following reasons.

First, real-world creative workflows are inherently iterative. Users typically generate multiple candidates at each editing step and select the most satisfactory result before proceeding to the next modification. This interactive process cannot be replicated by a single-pass prompt. Second, many practical scenarios involve secondary creative work on existing images, \eg, refining a previously generated AIGC result, which is by nature a multi-turn process. Third, prior works have established multi-turn image editing as an independent research challenge~\cite{MTC, qu2025vincie}, and the FLUX.1 Kontext technical report~\cite{labs2025flux1kontextflowmatching} explicitly identifies multi-turn editing as a key capability to develop.

From a technical standpoint, combining multiple editing instructions into a single prompt substantially increases the semantic complexity of the generation task, making it difficult for the model to disentangle and faithfully execute each individual operation while simultaneously preserving subject identity. As shown in Figure~\ref{fig:combined_vs_multiturn}, combining instructions often leads to significant subject identity drift: in the top row, the subject's facial appearance and surrounding background change substantially; in the bottom row, the left subject (red box) undergoes unintended appearance alterations. In contrast, our multi-turn editing approach applies instructions incrementally, enabling the model to maintain consistent subject identity across all editing turns while accurately executing each individual instruction.

\section{More Discussion on Limitations and Future Work}
\label{sec:discussion_limitation}

\vspace{5pt}
\noindent\textbf{Dependency on source image quality.}
FreqEdit relies on the high-frequency details present in the context image to compensate for progressive degradation across editing turns. When the initial source image is of low quality or already lacks fine-grained details, the high-frequency components available for injection are inherently limited, which constrains the effectiveness of our preservation mechanism. This limitation is fundamental to any reference-based injection approach and is not specific to our method.

\vspace{5pt}
\noindent\textbf{Out-of-distribution behavior.}
Our method modifies the velocity field at inference time without any model fine-tuning, which may steer the denoising trajectory into regions that lie outside the model's training distribution. The adaptive injection strategy (Section~\ref{sec:adaptive}) and path compensation mechanism (Section~\ref{sec:path_compensation}) are specifically designed to mitigate such deviations by spatially modulating injection strength, periodically realigning the trajectory with the editing objective. Nevertheless, these mechanisms reduce rather than eliminate the risk of out-of-distribution behavior, and edge cases may still arise for inputs with unusual appearance or highly atypical editing instructions.

\vspace{5pt}
\noindent\textbf{Reduced effectiveness for large-area edits.}
The adaptive injection map derives spatial injection weights from the divergence between the editing and reference velocity fields, which provides reliable region discrimination when edits are localized. When a single instruction targets a large portion of the image, \eg, a global style transfer, the divergence map loses discriminative power and the injection strength becomes less spatially precise. We note, however, that multi-turn editing workflows typically decompose complex modifications into a sequence of incremental changes, each affecting a limited region, which naturally mitigates this limitation in practice.

\vspace{5pt}
\noindent\textbf{Potential for training-based extensions.}
As a training-free framework, FreqEdit may not be optimized for specific editing types or distributions. The high-frequency preservation principles introduced in this work may nonetheless serve as a useful foundation for future training-based approaches, for instance by incorporating a frequency-aware consistency objective into the training loss, or by using our method to generate high-quality multi-turn training pairs for supervised fine-tuning.

\section{Multi-Turn Editing Instruction Generation}
\label{sec:appendix_templates}

We design a comprehensive instruction template (see Figure~\ref{fig:vlm_instruction}) to guide a VLM in generating structured image editing instructions. The template instructs the VLM to: (1) generate a detailed initial description capturing attributes such as color, shape, material, and spatial arrangement; (2) produce exactly 10 semantically coherent editing instructions that simulate a realistic editing workflow.

Each editing instruction is constrained to execute one primary operation from nine predefined categories: \textit{Subject Addition}, \textit{Subject Removal}, \textit{Subject Replacement}, \textit{Background Change}, \textit{Portrait Beautification}, \textit{Color Alteration}, \textit{Material Modification}, \textit{Motion Change}, and \textit{Style Transfer}. To ensure editability while preserving image identity, we enforce strict syntax rules: (i) mandatory preservation clauses specifying unchanged attributes, (ii) prohibition of ambiguous pronouns requiring precise descriptive references, (iii) a single-subject preservation rule preventing direct modifications to solitary primary objects.

To enable fair comparison with text-to-image editing baselines that require explicit source and target descriptions rather than editing instructions (\textit{e.g.}, inversion-based methods like MTC~\cite{MTC}), we design a secondary VLM instruction template for sequential description transformation (see Figure~\ref{fig:sequential_transformation}). Given the initial image description and the sequence of 10 editing instructions generated by the previous template, this instruction guides the VLM to produce 11 complete, standalone textual descriptions through iterative application. Starting from the initial description ($\text{description}[0]$), each editing instruction $p^{[i]}$ is applied sequentially to the previous description ($\text{description}[{i-1]}$) to generate the next state ($\text{description}[i]$). Crucially, each output description is self-contained and cumulative, incorporating all modifications from $p^{[1]}$ through $p^{[i]}$ while preserving unmodified attributes from the previous state. The resulting sequence enables the construction of 10 consecutive source-target pairs: $(\text{descriptions}[i], \text{descriptions}[i+1])$ for $i = 0$ to $9$, providing inversion-based editing models with the paired text descriptions required for their inference pipeline.

\vspace{5pt}
\noindent\textbf{Example of Generated Editing Instructions.}
We provide a representative example of the structured editing instructions generated by our VLM template. Given an input image depicting a young girl with a white dog on a leash in front of a storefront (see Figure~\ref{fig:qual_exp_2}), the VLM produces the following sequence of 10 editing prompts:

\begin{enumerate}
    \item\textit{The young girl with long dark hair is now raising her hand in a friendly wave towards the camera, while maintaining her same facial expression, clothing, and position.}
    
    \item\textit{The bright red leash is now a vibrant royal blue color, preserving the appearance of the young girl with long dark hair and the fluffy white dog.}
    
    \item\textit{The young girl's white platform sneakers are now made of a shiny, silver glitter material, keeping the same shape and style of the shoes.}
    
    \item\textit{A luxury paper shopping bag with rope handles, in a pastel pink color, is now placed on the sidewalk next to the young girl, preserving the appearance of the young girl with long dark hair and the fluffy white dog.}
    
    \item\textit{The patterns on the dress worn by the young girl with long dark hair are changed; the top is now solid light yellow, and the bottom tier has a blue and white polka dot pattern, preserving the girl's identity and the style of the dress.}
    
    \item\textit{Change the background from a modern storefront to the exterior of a charming Parisian café with a bistro table and chairs visible through the window, keeping the exact same camera angle, position, and framing, and preserving the girl and the dog.}
    
    \item\textit{Enhance the hair of the young girl with long dark hair to be more voluminous and styled with soft, flowing waves, while preserving her facial features and clothing.}
    
    \item\textit{Replace the white circular pouch at the girl's waist with a small, rectangular brown leather crossbody bag, keeping the rest of her outfit and appearance the same.}
    
    \item\textit{The young girl with long dark hair is now gently patting the top of the fluffy white dog's head with her hand, keeping her smiling expression and overall pose.}
    
    \item\textit{Transform the entire image's aesthetic into a bright, colorful watercolor painting style, preserving the recognizable features of the girl, the dog, and the café background.}
\end{enumerate}

\input{fig_tex/appendix_qualitative_gallery}

\input{fig_tex/instruction_generation}

\input{fig_tex/instruction_transformation}

%% file: tab_tex/additional_quantitative_average.tex
\begin{table*}[!t]
  \centering
  \small
  \setlength{\tabcolsep}{4pt}
  \caption{\textbf{Additional quantitative results using PSNR, SSIM and DINO-Sim.} 
  We report cumulative averages computed from turn 1 through each specified turn (1, 4, 7, 10) across 10 sequential edits. Bold indicates the best results, and underlined values denote the second-best results.}
  \resizebox{\textwidth}{!}{
  \begin{tabular}{
    l
    *{3}{c}
    *{3}{c}
    *{3}{c}
    *{3}{c}
  }
  \toprule
  \multirow{2}{*}{\textbf{Methods}} &
  \multicolumn{3}{c}{\textbf{Turn 1}} &
  \multicolumn{3}{c}{\textbf{Turn 4}} &
  \multicolumn{3}{c}{\textbf{Turn 7}} &
  \multicolumn{3}{c}{\textbf{Turn 10}} \\
  \cmidrule(lr){2-4}\cmidrule(lr){5-7}\cmidrule(lr){8-10}\cmidrule(lr){11-13}
  & \textbf{\makecell[c]{PSNR$\uparrow$}} 
  & \textbf{\makecell[c]{SSIM$\uparrow$}} 
  & \textbf{\makecell[c]{DINO-Sim$\uparrow$}} 
  & \textbf{\makecell[c]{PSNR$\uparrow$}} 
  & \textbf{\makecell[c]{SSIM$\uparrow$}} 
  & \textbf{\makecell[c]{DINO-Sim$\uparrow$}} 
  & \textbf{\makecell[c]{PSNR$\uparrow$}} 
  & \textbf{\makecell[c]{SSIM$\uparrow$}} 
  & \textbf{\makecell[c]{DINO-Sim$\uparrow$}} 
  & \textbf{\makecell[c]{PSNR$\uparrow$}} 
  & \textbf{\makecell[c]{SSIM$\uparrow$}} 
  & \textbf{\makecell[c]{DINO-Sim$\uparrow$}} \\
  \midrule
  Bagel 
  & 21.615 & 0.839 & 0.938 
  & 16.953 & 0.672 & 0.850 
  & 14.216 & 0.544 & 0.769 
  & 12.505 & 0.454 & 0.695 \\

  MTC 
  & 17.346 & 0.528 & 0.860 
  & 16.139 & 0.483 & 0.835 
  & \underline{15.618} & 0.466 & 0.816 
  & \textbf{15.291} & 0.456 & 0.799 \\

    Nano Banana 
  & 20.243 & 0.662 & \underline{0.963} 
  & 16.096 & 0.512 & \textbf{0.924} 
  & 14.210 & 0.446 & \underline{0.873} 
  & 12.983 & 0.403 & \underline{0.816} \\

  Seedream 4.0 
  & 19.885 & 0.737 & 0.957 
  & 16.916 & 0.603 & 0.878 
  & 14.749 & 0.514 & 0.790 
  & 13.259 & 0.458 & 0.706 \\

  VINCIE 
  & 17.648 & 0.687 & 0.881 
  & 14.766 & 0.587 & 0.790 
  & 13.095 & 0.510 & 0.717 
  & 12.121 & 0.459 & 0.658 \\

  \rowcolor{kontextblue}
  FLUX.1 Kontext + FreqEdit 
  & \cellcolor{kontextblueDeep}\underline{21.886} & \cellcolor{kontextblueDeep}\underline{0.844} & \cellcolor{kontextblueDeep}0.961
  & \cellcolor{kontextblueDeep}\underline{17.649} & \cellcolor{kontextblueDeep}\underline{0.707} & \cellcolor{kontextblueDeep}0.906
  & \cellcolor{kontextblueDeep}15.105 & \cellcolor{kontextblueDeep}\underline{0.591} & \cellcolor{kontextblueDeep}0.850
  & \cellcolor{kontextblueDeep}13.519 & \cellcolor{kontextblueDeep}\underline{0.514} & \cellcolor{kontextblueDeep}0.792 \\

  \rowcolor{kontextblue}
  FLUX.1 Kontext 
  & 18.998 & 0.724 & 0.953 
  & 15.214 & 0.597 & 0.884 
  & 13.228 & 0.510 & 0.817 
  & 11.942 & 0.446 & 0.746 \\

  \rowcolor{qwenred}
  Qwen-Image + FreqEdit 
  & \cellcolor{qwenredDeep}\textbf{23.548} & \cellcolor{qwenredDeep}\textbf{0.892} & 0.962
  & \cellcolor{qwenredDeep}\textbf{19.349} & \cellcolor{qwenredDeep}\textbf{0.798} & \cellcolor{qwenredDeep}\underline{0.920}
  & \cellcolor{qwenredDeep}\textbf{16.577} & \cellcolor{qwenredDeep}\textbf{0.706} & \cellcolor{qwenredDeep}\textbf{0.877}
  & \cellcolor{qwenredDeep}\underline{14.791} & \cellcolor{qwenredDeep}\textbf{0.632} & \cellcolor{qwenredDeep}\textbf{0.823} \\

  \rowcolor{qwenred}
  Qwen-Image 
  & 17.765 & 0.713 & \cellcolor{qwenredDeep}\textbf{0.965} 
  & 14.397 & 0.602 & 0.909 
  & 12.737 & 0.533 & 0.845 
  & 11.694 & 0.485 & 0.784 \\
  \bottomrule
  \end{tabular}
  }
  \label{tab:additional_quantitative}
\end{table*}

%% file: fig_tex/additional_quantitative_fig.tex
\begin{figure*}[htb]
  \centering
  \includegraphics[width=\textwidth]{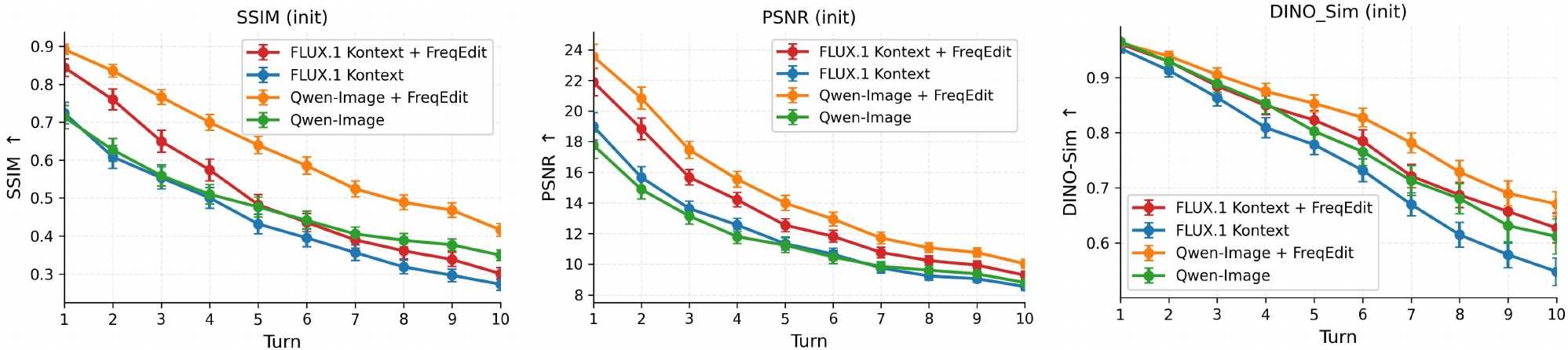}
  \caption{\textbf{Per-turn metrics across 10 sequential editing steps.} We report SSIM (left), PSNR (middle), and DINO-Sim (right) at each turn $k$, all computed by comparing the edited image $X^{[k+1]}$ with the original image $X^{[1]}$. For clarity, we only show our FreqEdit-enhanced models and their corresponding base models. FreqEdit consistently improves preservation of both low-level details (PSNR, SSIM) and high-level semantics (DINO-Sim) relative to the original image across all editing turns for both base models.}
  \label{fig:additional_quantitative_fig}
\end{figure*}

%% file: fig_tex/additional_ablation_AI.tex
\begin{figure}[tb]
    \centering
    \vspace{-8pt}
    \includegraphics[width=\linewidth]{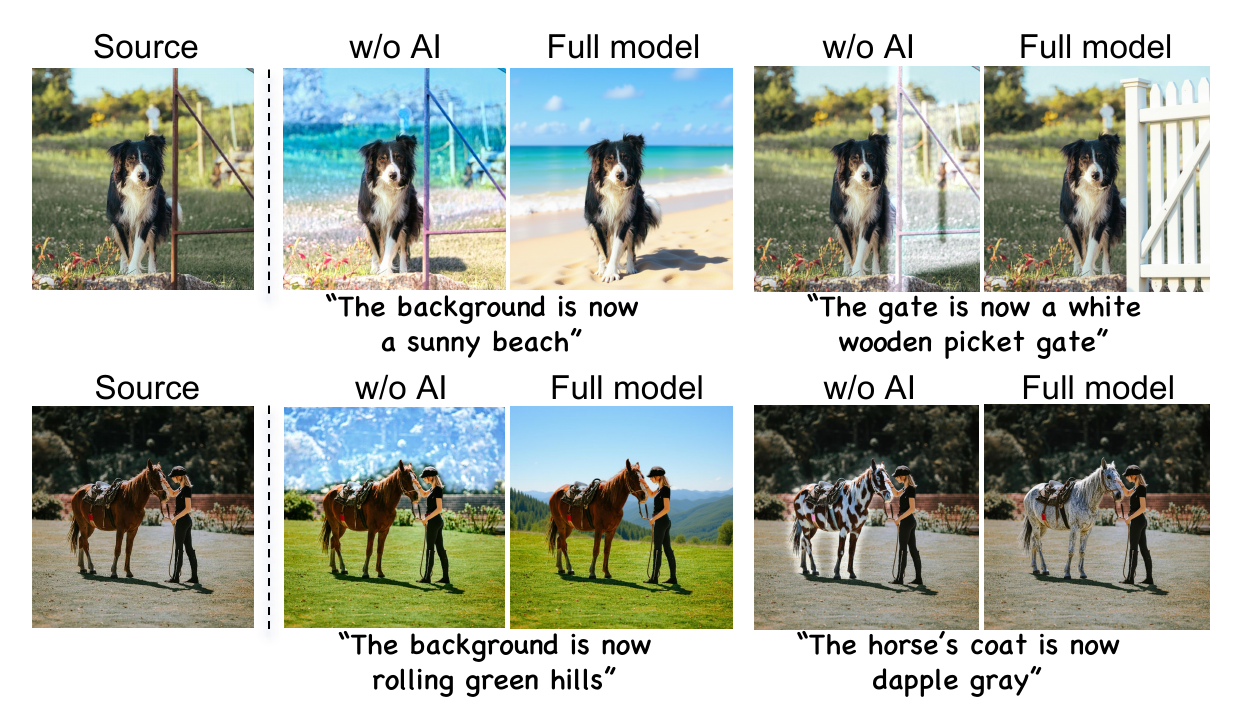}
    \caption{\textbf{Additional ablation results for Adaptive Injection (AI).} 
    The adaptive injection strategy modulates injection strength based on semantic correspondence between editing and reference velocity fields.
    When using uniform injection (w/o AI), semantically modified regions suffer from over-preservation, resulting in incomplete transformations for beach backgrounds, white gates, rolling hills, and horse coat patterns.
    }
    \label{fig:additional_ablation_AI}
\end{figure}

%% file: fig_tex/additional_ablation_PC.tex
\begin{figure}[tb]
    \centering
    \includegraphics[width=\linewidth]{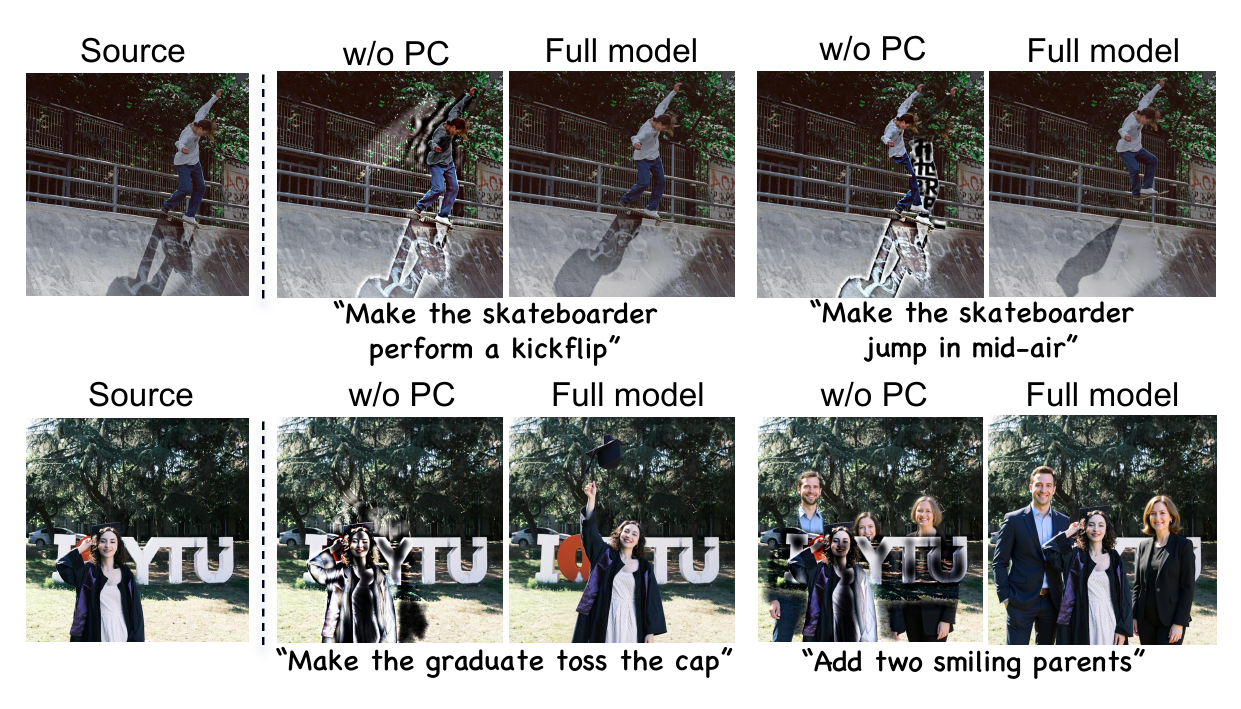}
    \caption{\textbf{Additional ablation results for Path Compensation (PC).} 
    Without PC, high injection strength introduces ghosting artifacts where conflicting visual elements from both editing and reference velocity fields manifest simultaneously (e.g., duplicated skateboarder and graduate poses, and ghost-like parent figures). 
    Our path compensation strategy eliminates these artifacts while maintaining strong high-frequency injection.}
    \label{fig:additional_ablation_PC}
\end{figure}

%% file: fig_tex/additional_ablation_QG.tex
\begin{figure}[tb]
    \centering
    \includegraphics[width=\linewidth]{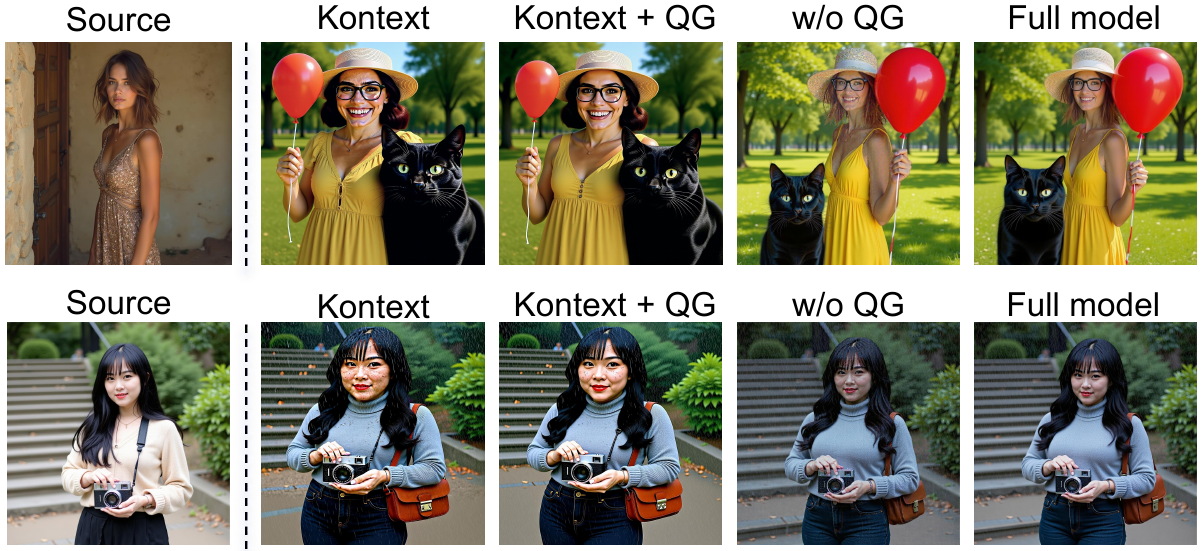}
    \caption{\textbf{Additional ablation results for Quality Guidance (QG) on FLUX.1 Kontext.} 
    After several editing iterations, noise artifacts accumulate progressively in the generated images. The native model exhibits severe noise degradation (column 2), while adding QG to the native model effectively suppresses noise accumulation (column 3). When combined with our wavelet-based injection framework, QG further enhances visual fidelity by eliminating residual noise artifacts (comparing columns 4 and 5).
    }
    \vspace{-8pt}
    \label{fig:additional_ablation_QG}
\end{figure}

%% file: fig_tex/additional_ablation_hyperparameters.tex
\begin{figure}[tb]
    \centering
    \includegraphics[width=\linewidth]{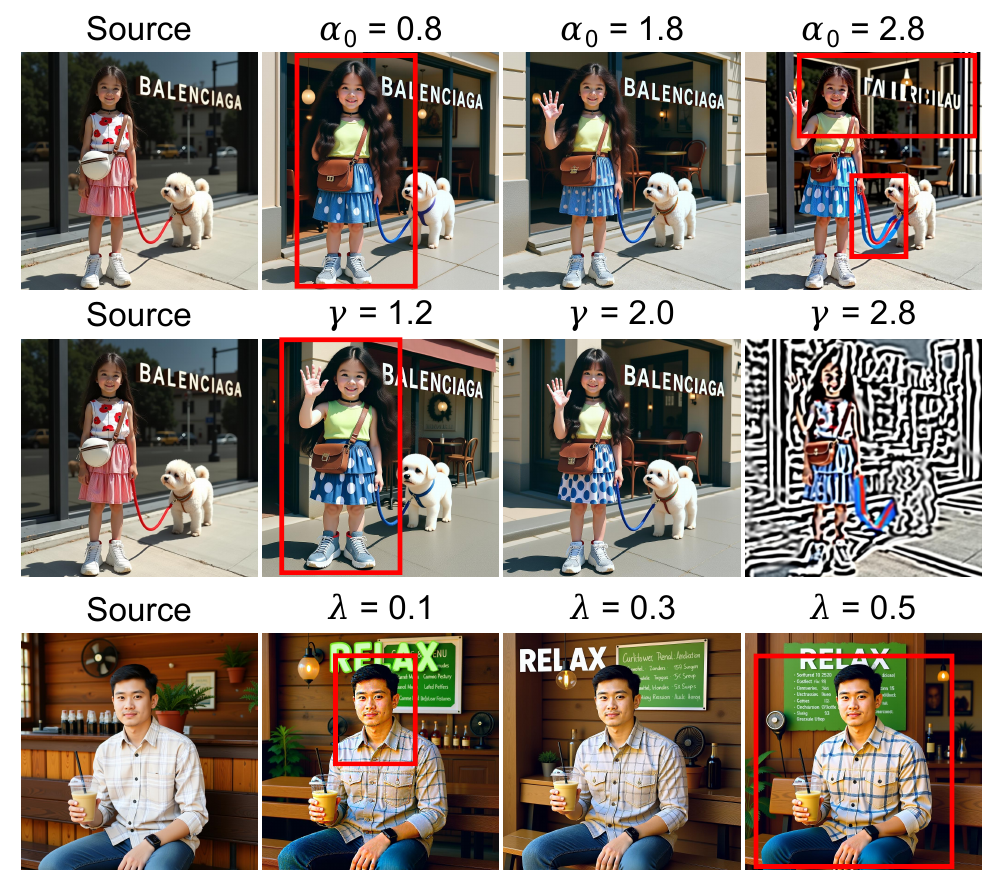}
    \caption{\textbf{Hyperparameter sensitivity analysis on FLUX.1 Kontext.}
    Each row varies one hyperparameter while fixing the others to their default values ($\alpha_0 = 1.6$, $\gamma = 2.0$, $\lambda = 0.3$); red boxes highlight failure regions. Insufficient $\alpha_0$ leads to subject deformation, while excessive $\alpha_0$ causes editing failure due to reference velocity dominance. $\gamma$ jointly controls injection strength and spatial map contrast: too small causes subject deformation; too large over-amplifies high-frequency components, corrupting image structure. For $\lambda$, too small leaves residual noise artifacts, while too large introduces over-saturation.}
    \label{fig:ablation_hyperparameters}
\end{figure}

%% file: fig_tex/frequency_alignment.tex
\begin{figure}[tb]
    \centering
    \includegraphics[width=\linewidth]{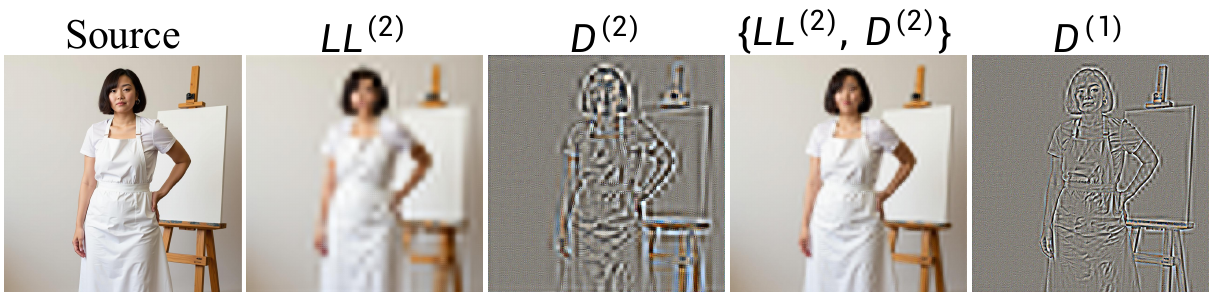}
    \caption{\textbf{Latent-space frequency alignment.}
    Images are reconstructed by retaining only the specified frequency band(s) from the 2-level DWT of the latent representation and decoding via VAE. $LL^{(2)}$ captures global structure and color; $D^{(2)}$ encodes coarser-scale edges and textures; $\{LL^{(2)}, D^{(2)}\}$ together yield a sharp reconstruction approximating the full image; $D^{(1)}$ captures the finest details such as hair strands and fabric textures. The consistent correspondence confirms that latent-space DWT reliably reflects image-space frequency structure.}
    \label{fig:frequency_alignment}
\end{figure}

%% file: fig_tex/visualization_PC.tex
\begin{figure}[tb]
    \centering
    \includegraphics[width=\linewidth]{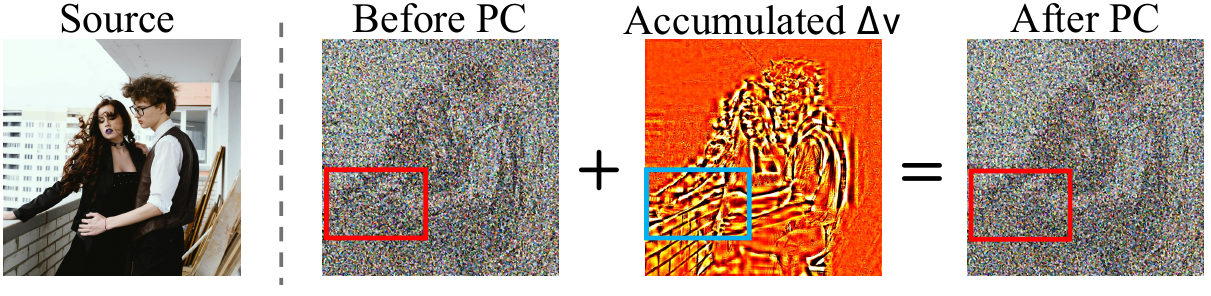}
    \caption{\textbf{Visualization of path compensation at an intermediate denoising step.}
    The corrected latent $Z_{t_i}$ after high-frequency injection exhibits residual structure conflicting with the editing objective (red box). The trajectory buffer ${B} = \sum (t_{i-1} - t_i) \cdot \Delta v_{t_i}$, visualized as a heatmap, carries a structured correction signal in the corresponding region (blue box) that counteracts this residual structure. Adding $\mathcal{B}$ to the latent yields a clean latent aligned with the intended edit. Please zoom in for details.}
    \label{fig:visualization_PC}
\end{figure}

%% file: fig_tex/combined_vs_multiturn.tex
\begin{figure}[tb]
    \centering
    \includegraphics[width=\linewidth]{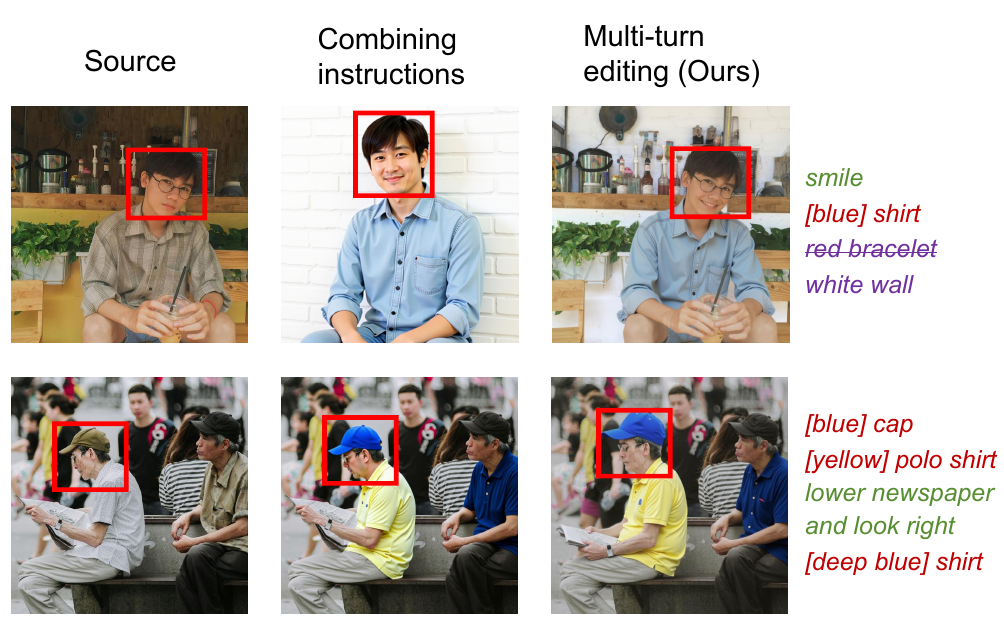}
    \caption{\textbf{Comparison between combining instructions and multi-turn editing.}
    Red boxes highlight subject regions for identity comparison. Combining all instructions into a single prompt causes unintended subject identity drift and background changes, despite these regions not being targeted by any instruction. Our multi-turn approach applies instructions incrementally, preserving subject identity while accurately executing each individual edit.}
    \label{fig:combined_vs_multiturn}
\end{figure}

%% file: fig_tex/appendix_qualitative_gallery.tex
\begin{figure*}[p] 
    \centering
    \includegraphics[width=\linewidth, height=0.92\textheight, keepaspectratio]{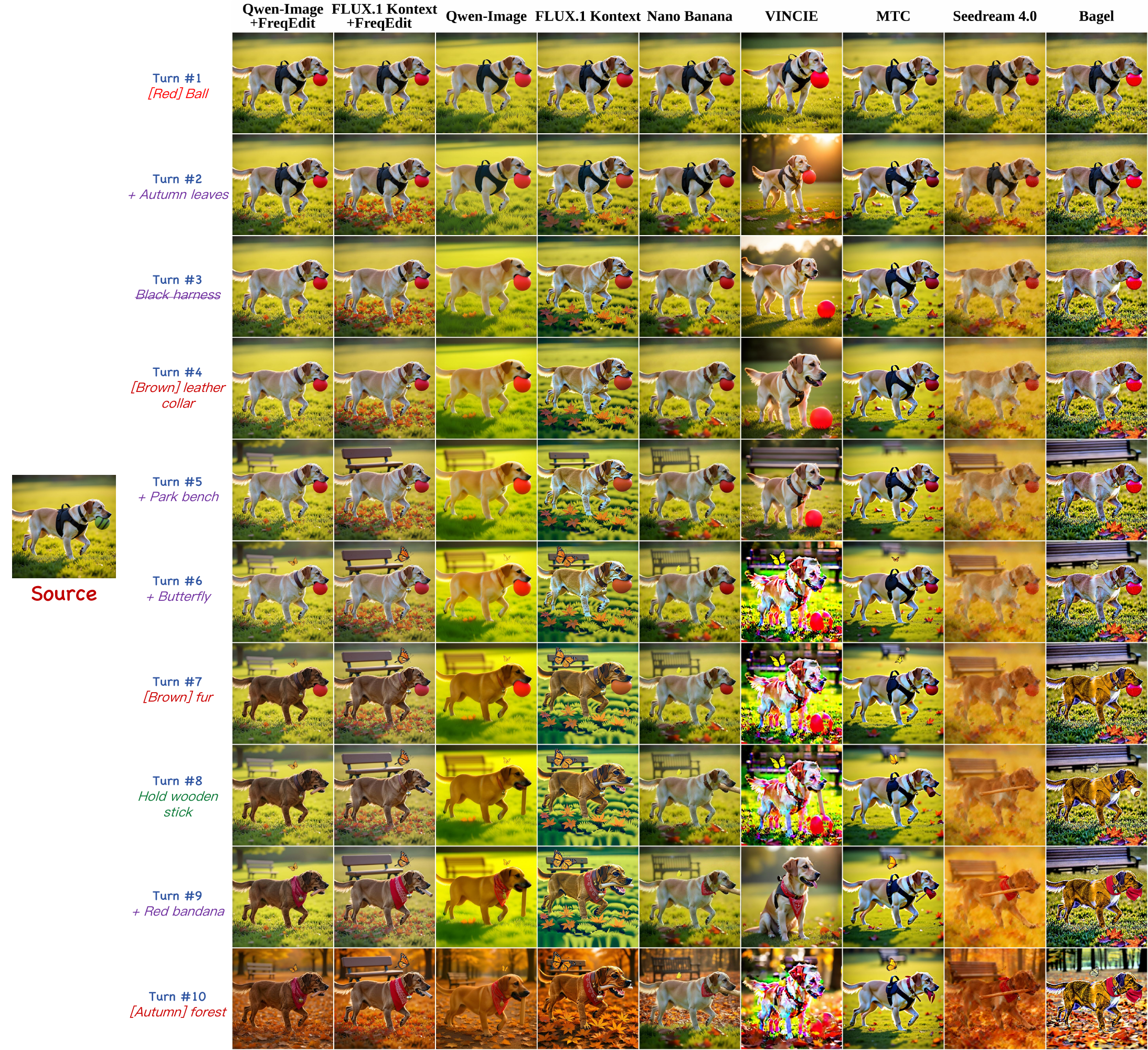}
    \caption{\textbf{Additional qualitative comparison (1/6).} We compare our method against several state-of-the-art methods, including FLUX.1 Kontext~\cite{labs2025flux1kontextflowmatching}, Qwen-Image~\cite{wu2025qwenimagetechnicalreport}, Seedream 4.0~\cite{seedream2025seedream40nextgenerationmultimodal}, Nano Banana~\cite{comanici2025gemini25pushingfrontier}, MTC~\cite{MTC}, VINCIE~\cite{qu2025vincie}, and Bagel~\cite{deng2025bagel}.
    Please zoom in for a better view.}
    \label{fig:qual_exp_1}
\end{figure*}
\clearpage 

\begin{figure*}[p]
    \centering
    \includegraphics[width=\linewidth, height=0.92\textheight, keepaspectratio]{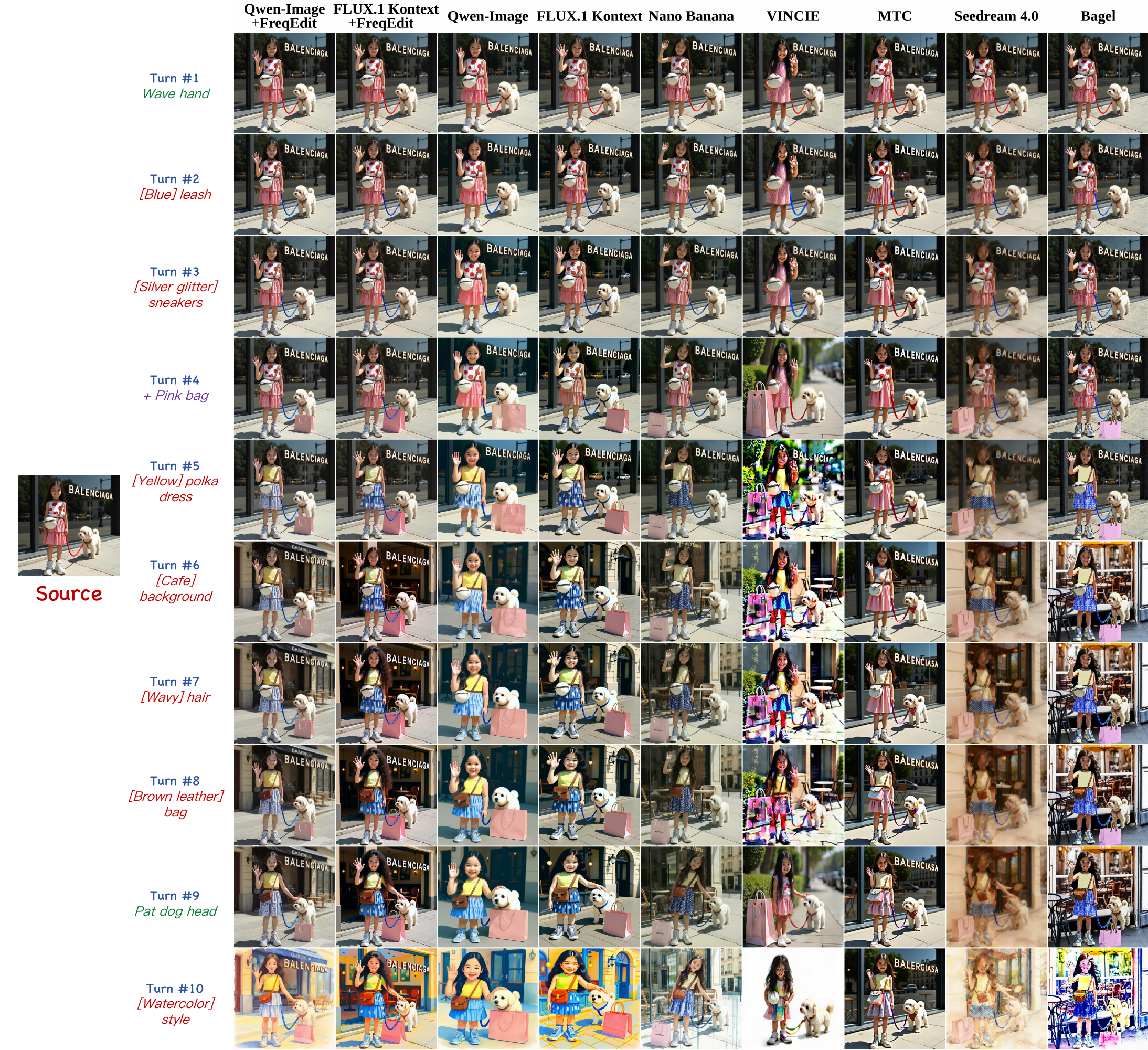}
    \caption{\textbf{Additional qualitative comparison (2/6).} We compare our method against several state-of-the-art methods, including FLUX.1 Kontext~\cite{labs2025flux1kontextflowmatching}, Qwen-Image~\cite{wu2025qwenimagetechnicalreport}, Seedream 4.0~\cite{seedream2025seedream40nextgenerationmultimodal}, Nano Banana~\cite{comanici2025gemini25pushingfrontier}, MTC~\cite{MTC}, VINCIE~\cite{qu2025vincie}, and Bagel~\cite{deng2025bagel}.
    Please zoom in for a better view.}
    \label{fig:qual_exp_2}
\end{figure*}
\clearpage

\begin{figure*}[p]
    \centering
    \includegraphics[width=\linewidth, height=0.92\textheight, keepaspectratio]{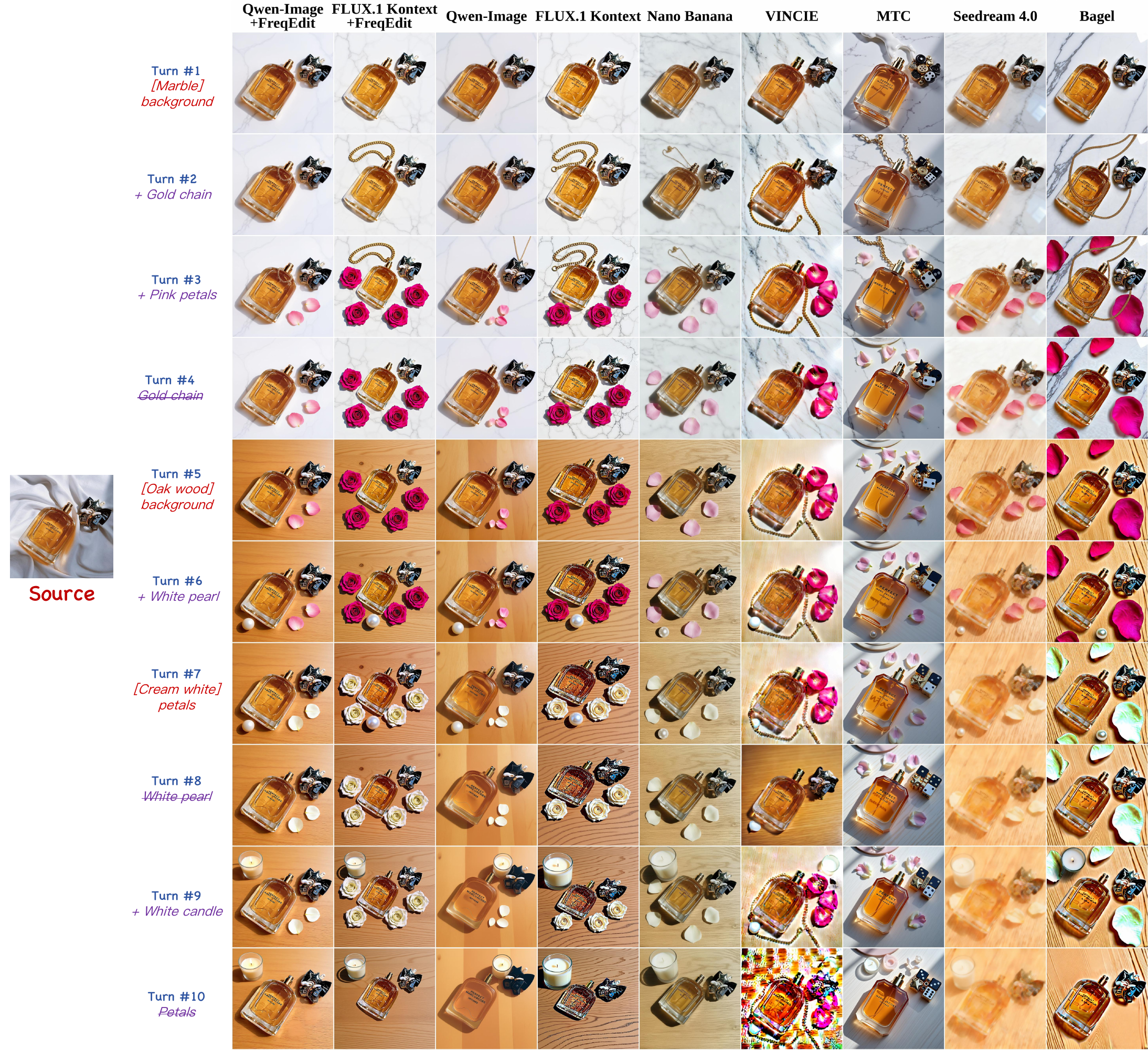}
    \caption{\textbf{Additional qualitative comparison (3/6).} We compare our method against several state-of-the-art methods, including FLUX.1 Kontext~\cite{labs2025flux1kontextflowmatching}, Qwen-Image~\cite{wu2025qwenimagetechnicalreport}, Seedream 4.0~\cite{seedream2025seedream40nextgenerationmultimodal}, Nano Banana~\cite{comanici2025gemini25pushingfrontier}, MTC~\cite{MTC}, VINCIE~\cite{qu2025vincie}, and Bagel~\cite{deng2025bagel}.
    Please zoom in for a better view.}
    \label{fig:qual_exp_3}
\end{figure*}
\clearpage

\begin{figure*}[p]
    \centering
    \includegraphics[width=\linewidth, height=0.92\textheight, keepaspectratio]{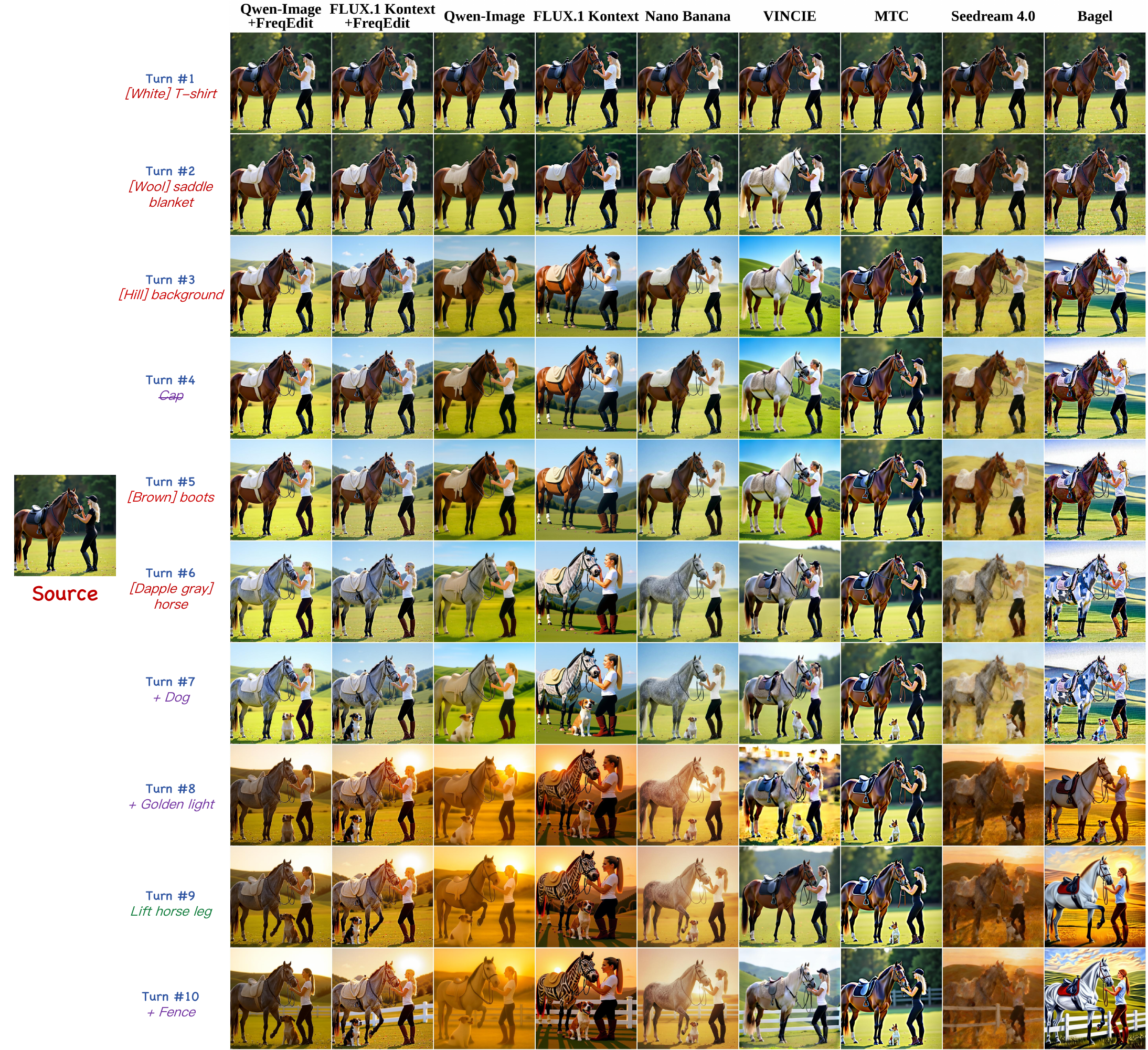}
    \caption{\textbf{Additional qualitative comparison (4/6).} We compare our method against several state-of-the-art methods, including FLUX.1 Kontext~\cite{labs2025flux1kontextflowmatching}, Qwen-Image~\cite{wu2025qwenimagetechnicalreport}, Seedream 4.0~\cite{seedream2025seedream40nextgenerationmultimodal}, Nano Banana~\cite{comanici2025gemini25pushingfrontier}, MTC~\cite{MTC}, VINCIE~\cite{qu2025vincie}, and Bagel~\cite{deng2025bagel}.
    Please zoom in for a better view.}
    \label{fig:qual_exp_4}
\end{figure*}
\clearpage

\begin{figure*}[p]
    \centering
    \includegraphics[width=\linewidth, height=0.92\textheight, keepaspectratio]{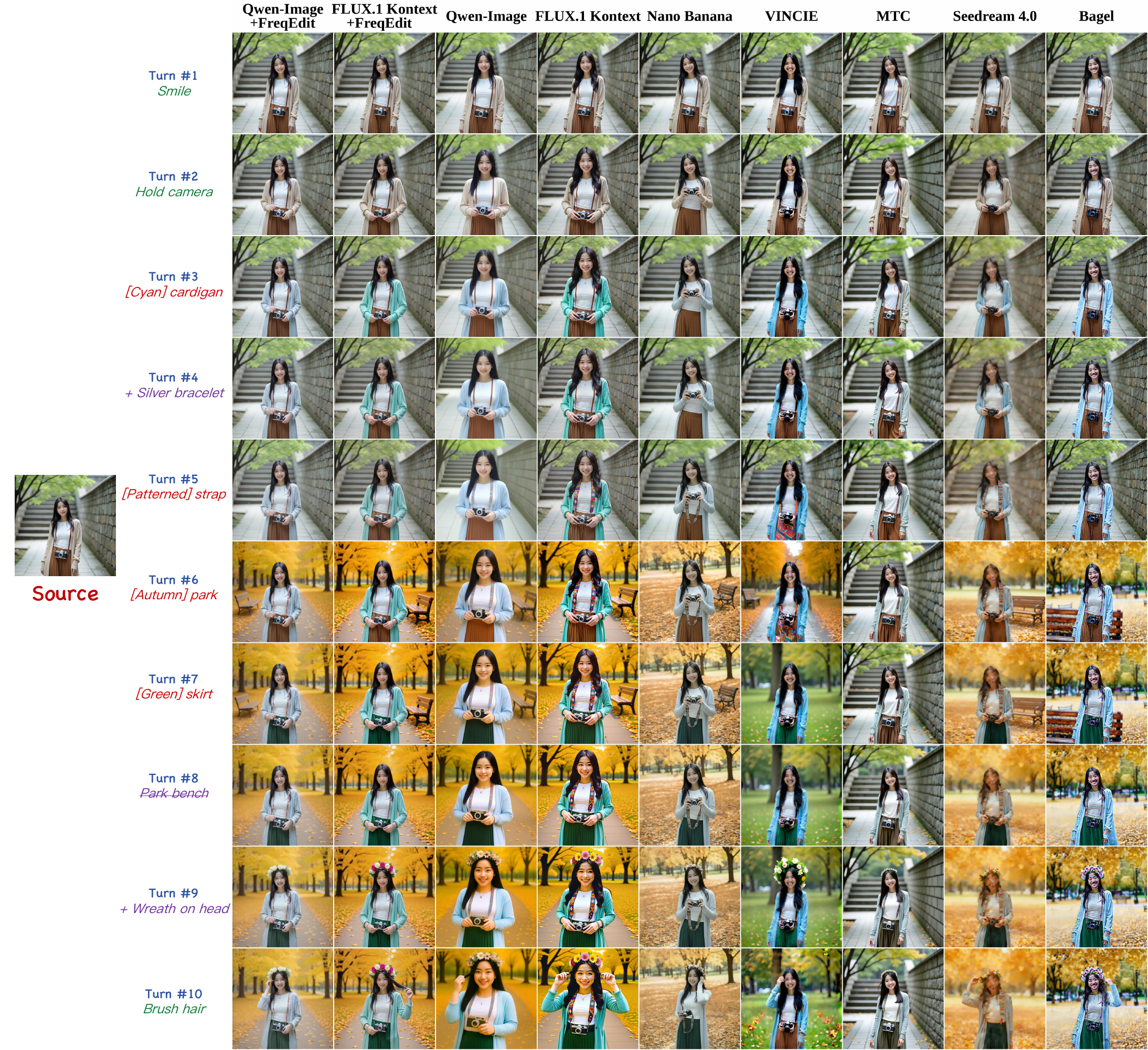}
    \caption{\textbf{Additional qualitative comparison (5/6).} We compare our method against several state-of-the-art methods, including FLUX.1 Kontext~\cite{labs2025flux1kontextflowmatching}, Qwen-Image~\cite{wu2025qwenimagetechnicalreport}, Seedream 4.0~\cite{seedream2025seedream40nextgenerationmultimodal}, Nano Banana~\cite{comanici2025gemini25pushingfrontier}, MTC~\cite{MTC}, VINCIE~\cite{qu2025vincie}, and Bagel~\cite{deng2025bagel}.
    Please zoom in for a better view.}
    \label{fig:qual_exp_5}
\end{figure*}
\clearpage

\begin{figure*}[p]
    \centering
    \includegraphics[width=\linewidth, height=0.92\textheight, keepaspectratio]{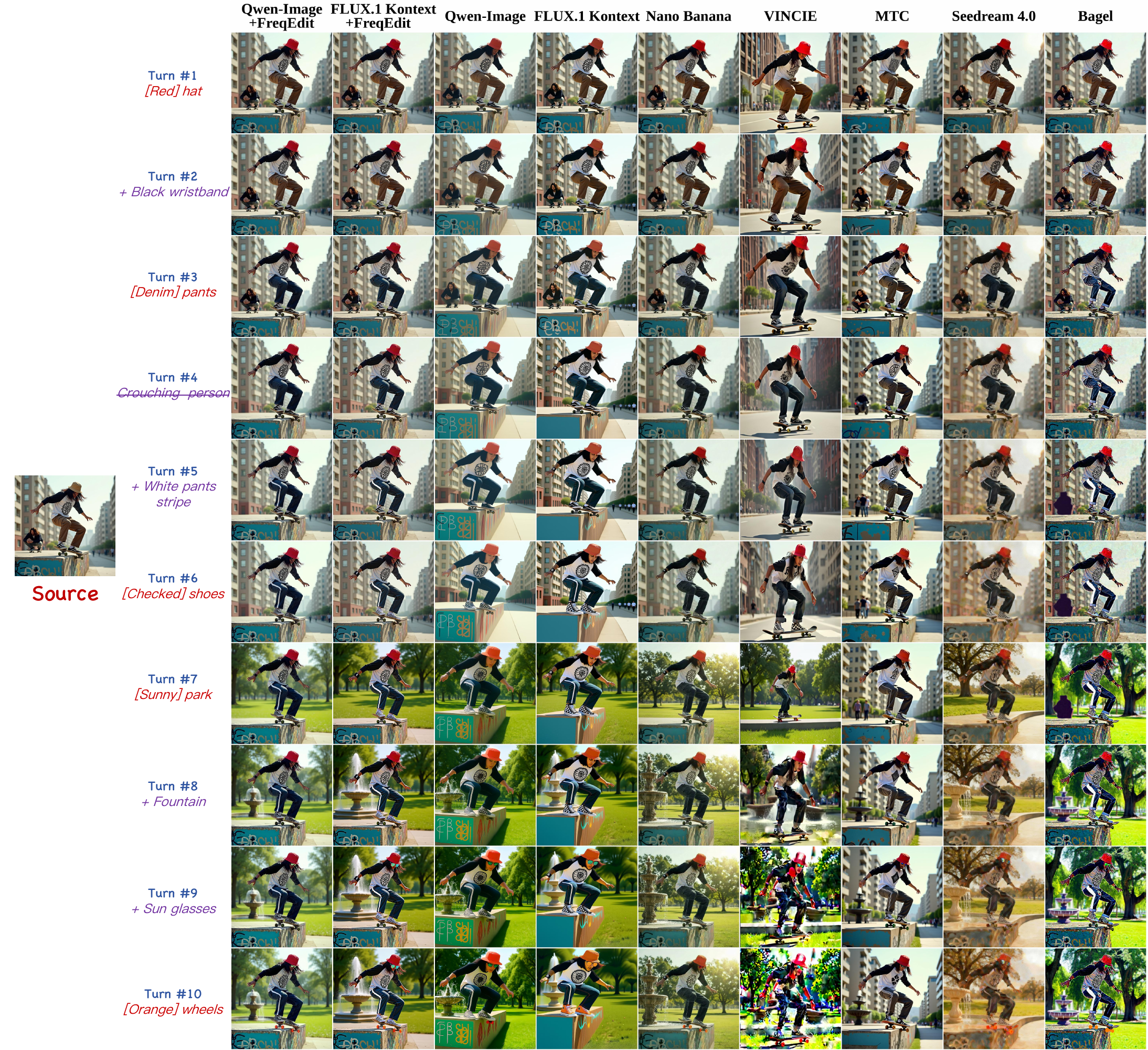}
    \caption{\textbf{Additional qualitative comparison (6/6).} We compare our method against several state-of-the-art methods, including FLUX.1 Kontext~\cite{labs2025flux1kontextflowmatching}, Qwen-Image~\cite{wu2025qwenimagetechnicalreport}, Seedream 4.0~\cite{seedream2025seedream40nextgenerationmultimodal}, Nano Banana~\cite{comanici2025gemini25pushingfrontier}, MTC~\cite{MTC}, VINCIE~\cite{qu2025vincie}, and Bagel~\cite{deng2025bagel}.
    Please zoom in for a better view.}
    \label{fig:qual_exp_6}
\end{figure*}
\clearpage

%% file: fig_tex/instruction_generation.tex
\begin{figure*}[b]
\centering

\begin{tcolorbox}[
    width=0.98\textwidth,
    colback=gray!20,
    colframe=gray!60,
    arc=3mm,
    boxrule=0.5pt,
    title=Structured Prompt Template for Multi-Turn Image Editing Instruction Generation,
    fonttitle=\bfseries,
    coltitle=black,
    colbacktitle=gray!40
]

\textbf{Imagine that you are a top-tier AI Prompt Engineer.} You are an expert in image analysis, JSON data structures, and a specific, rigorous English syntax for generating image editing prompts. Your output must adhere 100\% to the syntax and formatting rules defined in this directive.

\vspace{0.6em}
\noindent\textbf{Your task is to analyze a user-provided image} (or its textual description) and output a single, syntactically perfect JSON object. This object will contain an initial description of the image, a descriptive tag, and an array of 10 distinct English editing prompts that follow a strict, sequential logic.

\vspace{0.6em}
\noindent\textbf{You must generate the editing prompts from the following guidelines}:

\noindent\textbf{1.} Analyze the image first. Identify all key subjects, elements, the environment, and the composition. Assign specific, descriptive names to key subjects for subsequent reference (\textit{e.g.}, ``the woman with short black hair'', ``the snowy mountain on the left'').

\noindent\textbf{2.} Generate a comprehensive initial prompt (\texttt{init\_prompt}). For every significant object, subject, and background element, describe its key attributes including precise color, shape, material, texture, and position. Do not mention the real name of entities.

\noindent\textbf{3.} Generate a tag in the format \texttt{[Formatted\_Filename]\_[Subject]\_[Scene]}. Format numbers to three digits with leading zeros (\textit{e.g.}, input `1.png' results in `001').

\noindent\textbf{4.} Generate exactly 10 editing prompts. Each must be unique and adhere to the syntax rules below. Avoid random, disconnected edits.

\noindent\textbf{5.} The 10 prompts must be logical and thematically cohesive, simulating a real editor's workflow.

\vspace{0.6em}
\noindent\textbf{An editing prompt should follow these rules}:

\noindent\textbf{1.} Execute only one primary operation from: \texttt{Subject Addition}, \texttt{Subject Removal}, \texttt{Subject Replacement}, \texttt{Background Change}, \texttt{Portrait Beautification}, \texttt{Color Alteration}, \texttt{Material Modification}, \texttt{Motion Change}, \texttt{Style Transfer}.

\noindent\textbf{2.} Single Subject Preservation Rule: If the image contains only one primary subject (\textit{e.g.}, product photography), you are strictly forbidden from making any direct changes to that subject. Edits can modify the background, add/remove surrounding objects, and adjust lighting only.

\noindent\textbf{3.} Mandatory Preservation Clause: Every prompt must end with an explicit preservation clause using: \textit{while maintaining...}, \textit{preserving...}, or \textit{keeping...} (\textit{e.g.}, ``while maintaining her same facial features, hairstyle, and overall appearance'').

\noindent\textbf{4.} No Pronouns: Strictly forbidden from using ambiguous pronouns like he, she, it, they, this, or that. Always use specific descriptive names.

\noindent\textbf{5.} Extreme Specificity: Use precise descriptions (\textit{e.g.}, ``a bright scarlet red T-shirt'' instead of ``a red shirt'').

\noindent\textbf{6.} Compositional Control: When changing backgrounds, explicitly lock camera viewpoint: \textit{``...while keeping the exact same camera angle, position, and framing''}.

\noindent\textbf{7.} Style Constraint: You are prohibited from generating prompts that create a sci-fi or dark/gloomy aesthetic. The ``Style Transfer'' operations must only be placed between steps 7 and 10.

\vspace{0.6em}
\noindent\textbf{Output Format:} You should output a JSON file to include the following information:

\noindent\texttt{tag}: formatted filename with subject and scene

\noindent\texttt{init\_prompt}: comprehensive photorealistic description of the initial image

\noindent\texttt{editing\_prompts}: array of exactly 10 editing instructions following all syntax rules

\vspace{0.2em}
\noindent Your output should be a JSON file in one row (without any format), which looks like:

\noindent\texttt{\{"tag": "001\_Woman\_By\_Wall", "init\_prompt": "A full-body photo of a woman with short black hair standing in front of a brick wall, wearing a blue t-shirt and jeans.", "editing\_prompts": ["The woman's blue t-shirt is now a bright scarlet red, while maintaining her same facial features, hairstyle, and overall appearance.", ..., "The entire image is transformed into a Claymation style, while keeping the same person and the overall composition."]\}}

\end{tcolorbox}

\caption{\textbf{VLM prompt template for generating structured image editing instructions.}}
\vspace{-15pt}
\label{fig:vlm_instruction}
\end{figure*}

%% file: fig_tex/instruction_transformation.tex
\begin{figure*}[t]
\centering

\begin{tcolorbox}[
    width=0.98\textwidth,
    colback=gray!20,
    colframe=gray!60,
    arc=3mm,
    boxrule=0.5pt,
    title=Structured Prompt Template for Image Description Transformation,
    fonttitle=\bfseries,
    coltitle=black,
    colbacktitle=gray!40
]

\textbf{Imagine that you are a meticulous prompt engineer.} You should take an initial Base Description and a sequence of Editing Instructions, then generate a complete list of transformed image descriptions.

\vspace{0.6em}
\noindent\textbf{Your task is to transform the base description sequentially} by applying each editing instruction one at a time, where each output becomes the input for the next step.

\vspace{0.6em}
\noindent\textbf{You must follow this transformation workflow}:

\noindent\textbf{1.} Start with \texttt{description\_0} = \texttt{init\_prompt} (the initial base description).

\noindent\textbf{2.} For each editing instruction $p^{[i]}$ (where $i$ ranges from 1 to 10):

\quad $\bullet$ Take the previous description \texttt{description\_{i-1}} as the base.

\quad $\bullet$ Apply only the single editing instruction $p^{[i]}$ while preserving all unmodified details.

\quad $\bullet$ Generate \texttt{description\_i} as the new complete, standalone description.

\noindent\textbf{3.} Output a dictionary containing the tag and list of all 11 descriptions.

\vspace{0.6em}
\noindent\textbf{Key principles}:

\noindent$\bullet$ Each description must be complete and standalone (not incremental changes).

\noindent$\bullet$ Perfectly preserve all details from the previous step that are not modified by the current edit.

\noindent$\bullet$ Apply edits cumulatively: \texttt{description\_i} includes all changes from $p^{[1]}$ through $p^{[i]}$.

\vspace{0.6em}
\noindent\textbf{Input Format}:

\noindent\textbf{Tag:} \texttt{<tag>}

\noindent\textbf{Initial Description:} \texttt{<init\_prompt>}

\noindent\textbf{Editing Instructions (in order):}

\noindent 1. \texttt{<editing\_instruction\_1>}

\noindent 2. \texttt{<editing\_instruction\_2>}

\noindent ...

\noindent 10. \texttt{<editing\_instruction\_10>}

\vspace{0.6em}
\noindent\textbf{Example}:

\noindent\textbf{Input:}

\noindent Tag: ``001\_Woman\_By\_Wall''

\noindent Initial: ``A full-body photo of a woman with short black hair standing in front of a brick wall, wearing a blue t-shirt and jeans.''

\noindent Edits: [``t-shirt now red, maintaining...'', ``hair now wavy, preserving...'', ...]

\noindent\textbf{Output:}

\noindent\texttt{\{"tag": "001\_Woman\_By\_Wall",}

\noindent\texttt{\quad"descriptions": [}

\noindent\texttt{\quad\quad"A full-body photo of a woman with short black hair standing in front of a brick wall, wearing a blue t-shirt and jeans.",}

\noindent\texttt{\quad\quad"A full-body photo of a woman with short black hair standing in front of a brick wall, wearing a red t-shirt and jeans.",}

\noindent\texttt{\quad\quad"A full-body photo of a woman with wavy black hair standing in front of a brick wall, wearing a red t-shirt and jeans.",}

\noindent\texttt{\quad\quad...}

\noindent\texttt{\quad]\}}

\vspace{0.6em}
\noindent\textbf{Output Format:} A JSON object with tag and descriptions array:

\noindent\texttt{\{"tag": "<tag>", "descriptions": ["<description\_0>", "<description\_1>", ..., "<description\_10>"]\}}

\end{tcolorbox}

\caption{\textbf{VLM prompt template for sequential description transformation.}
The system applies all editing instructions in one pass, generating a dictionary containing the tag and a list of 11 descriptions (the initial state followed by 10 transformed states). Source–target pairs are then constructed as consecutive descriptions, i.e., (\texttt{descriptions[i]}, \texttt{descriptions[i+1]}) for $i=0$ to $9$, for use in inversion-based image editing models (e.g., MTC~\cite{MTC}).}

\label{fig:sequential_transformation}
\end{figure*}

%% file: main.bbl
\begin{thebibliography}{62}
\providecommand{\natexlab}[1]{#1}
\providecommand{\url}[1]{\texttt{#1}}
\expandafter\ifx\csname urlstyle\endcsname\relax
  \providecommand{\doi}[1]{doi: #1}\else
  \providecommand{\doi}{doi: \begingroup \urlstyle{rm}\Url}\fi

\bibitem[Avrahami et~al.(2022)Avrahami, Lischinski, and Fried]{Avrahami_2022}
Omri Avrahami, Dani Lischinski, and Ohad Fried.
\newblock Blended diffusion for text-driven editing of natural images.
\newblock In \emph{CVPR}, 2022.

\bibitem[Avrahami et~al.(2025)Avrahami, Patashnik, Fried, Nemchinov, Aberman, Lischinski, and Cohen-Or]{Avrahami_2025_CVPR}
Omri Avrahami, Or Patashnik, Ohad Fried, Egor Nemchinov, Kfir Aberman, Dani Lischinski, and Daniel Cohen-Or.
\newblock Stable flow: Vital layers for training-free image editing.
\newblock In \emph{CVPR}, 2025.

\bibitem[Brooks et~al.(2022)Brooks, Holynski, and Efros]{brooks2022instructpix2pix}
Tim Brooks, Aleksander Holynski, and Alexei~A Efros.
\newblock Instructpix2pix: Learning to follow image editing instructions.
\newblock \emph{arXiv preprint arXiv:2211.09800}, 2022.

\bibitem[Cao et~al.(2023)Cao, Wang, Qi, Shan, Qie, and Zheng]{cao_2023_masactrl}
Mingdeng Cao, Xintao Wang, Zhongang Qi, Ying Shan, Xiaohu Qie, and Yinqiang Zheng.
\newblock Masactrl: Tuning-free mutual self-attention control for consistent image synthesis and editing.
\newblock In \emph{ICCV}, 2023.

\bibitem[Chefer et~al.(2023)Chefer, Alaluf, Vinker, Wolf, and Cohen-Or]{chefer2023attendandexcite}
Hila Chefer, Yuval Alaluf, Yael Vinker, Lior Wolf, and Daniel Cohen-Or.
\newblock Attend-and-excite: Attention-based semantic guidance for text-to-image diffusion models.
\newblock \emph{arXiv preprint arXiv:2301.13826}, 2023.

\bibitem[Chen et~al.(2025)Chen, Zhang, Zhang, Yu, Wang, Wang, Lin, Wang, Yang, Li, Lin, Xie, Leong, Wang, Wu, and Zhou]{EdiVal}
Tianyu Chen, Yasi Zhang, Zhi Zhang, Peiyu Yu, Shu Wang, Zhendong Wang, Kevin Lin, Xiaofei Wang, Zhengyuan Yang, Linjie Li, Chung-Ching Lin, Jianwen Xie, Oscar Leong, Lijuan Wang, Ying~Nian Wu, and Mingyuan Zhou.
\newblock Edival-agent: An object-centric framework for automated, fine-grained evaluation of multi-turn editing.
\newblock \emph{arXiv preprint arXiv:2509.13399}, 2025.

\bibitem[Couairon et~al.(2022)Couairon, Verbeek, Schwenk, and Cord]{couairon2022diffeditdiffusionbasedsemanticimage}
Guillaume Couairon, Jakob Verbeek, Holger Schwenk, and Matthieu Cord.
\newblock Diffedit: Diffusion-based semantic image editing with mask guidance.
\newblock \emph{arXiv preprint arXiv:2210.11427}, 2022.

\bibitem[Daubechies(1992)]{daubechies1992ten}
I. Daubechies.
\newblock \emph{Ten Lectures on Wavelets}.
\newblock Society for Industrial and Applied Mathematics, 1992.

\bibitem[Deng et~al.(2025)Deng, Zhu, Li, Gou, Li, Wang, Zhong, Yu, Nie, Song, Shi, and Fan]{deng2025bagel}
Chaorui Deng, Deyao Zhu, Kunchang Li, Chenhui Gou, Feng Li, Zeyu Wang, Shu Zhong, Weihao Yu, Xiaonan Nie, Ziang Song, Guang Shi, and Haoqi Fan.
\newblock Emerging properties in unified multimodal pretraining.
\newblock \emph{arXiv preprint arXiv:2505.14683}, 2025.

\bibitem[Deng et~al.(2024)Deng, He, Mei, Wang, and Tang]{deng2024fireflowfastinversionrectified}
Yingying Deng, Xiangyu He, Changwang Mei, Peisong Wang, and Fan Tang.
\newblock Fireflow: Fast inversion of rectified flow for image semantic editing.
\newblock \emph{arXiv preprint arXiv:2412.07517}, 2024.

\bibitem[Esser et~al.(2024)Esser, Kulal, Blattmann, Entezari, Müller, Saini, Levi, Lorenz, Sauer, Boesel, Podell, Dockhorn, English, Lacey, Goodwin, Marek, and Rombach]{sd3}
Patrick Esser, Sumith Kulal, Andreas Blattmann, Rahim Entezari, Jonas Müller, Harry Saini, Yam Levi, Dominik Lorenz, Axel Sauer, Frederic Boesel, Dustin Podell, Tim Dockhorn, Zion English, Kyle Lacey, Alex Goodwin, Yannik Marek, and Robin Rombach.
\newblock Scaling rectified flow transformers for high-resolution image synthesis.
\newblock \emph{arXiv preprint arXiv:2403.03206}, 2024.

\bibitem[Fu et~al.(2024)Fu, Hu, Du, Wang, Yang, and Gan]{fu2024mgie}
Tsu-Jui Fu, Wenze Hu, Xianzhi Du, William~Yang Wang, Yinfei Yang, and Zhe Gan.
\newblock {Guiding Instruction-based Image Editing via Multimodal Large Language Models}.
\newblock In \emph{ICLR}, 2024.

\bibitem[Gao et~al.(2024)Gao, Xu, Zhao, and Liu]{gao2024frequency}
Xiang Gao, Zhengbo Xu, Junhan Zhao, and Jiaying Liu.
\newblock Frequency-controlled diffusion model for versatile text-guided image-to-image translation.
\newblock In \emph{AAAI}, 2024.

\bibitem[Ge et~al.(2024)Ge, Zhao, Li, Ge, and Shan]{ge2024seeddataedittechnicalreporthybrid}
Yuying Ge, Sijie Zhao, Chen Li, Yixiao Ge, and Ying Shan.
\newblock Seed-data-edit technical report: A hybrid dataset for instructional image editing.
\newblock \emph{arXiv preprint arXiv:2405.04007}, 2024.

\bibitem[Geng et~al.(2024)Geng, Yang, Hang, Li, Gu, Zhang, Bao, Zhang, Hu, Chen, and Guo]{Geng23instructdiff}
Zigang Geng, Binxin Yang, Tiankai Hang, Chen Li, Shuyang Gu, Ting Zhang, Jianmin Bao, Zheng Zhang, Han Hu, Dong Chen, and Baining Guo.
\newblock Instructdiffusion: {A} generalist modeling interface for vision tasks.
\newblock In \emph{CVPR}, 2024.

\bibitem[Geng et~al.(2025)Geng, Deng, Bai, Kolter, and He]{meanflow}
Zhengyang Geng, Mingyang Deng, Xingjian Bai, J.~Zico Kolter, and Kaiming He.
\newblock Mean flows for one-step generative modeling.
\newblock \emph{arXiv preprint arXiv:2505.13447}, 2025.

\bibitem[Gonzalez and Woods(2008)]{gonzalez2008digital}
R.C. Gonzalez and R.E. Woods.
\newblock \emph{Digital Image Processing}.
\newblock Prentice Hall, 2008.

\bibitem[{Google DeepMind et al.}(2025)]{comanici2025gemini25pushingfrontier}
{Google DeepMind et al.}
\newblock Gemini 2.5: Pushing the frontier with advanced reasoning, multimodality, long context, and next generation agentic capabilities.
\newblock \emph{arXiv preprint arXiv:2507.06261}, 2025.

\bibitem[Guo and Lin(2024)]{guo2023focusinstructionfinegrainedmultiinstruction}
Qin Guo and Tianwei Lin.
\newblock Focus on your instruction: Fine-grained and multi-instruction image editing by attention modulation.
\newblock In \emph{CVPR}, 2024.

\bibitem[Han et~al.(2024)Han, Jiang, Pan, Zhang, Mao, Xie, Liu, and Zhou]{ACE}
Zhen Han, Zeyinzi Jiang, Yulin Pan, Jingfeng Zhang, Chaojie Mao, Chenwei Xie, Yu Liu, and Jingren Zhou.
\newblock Ace: All-round creator and editor following instructions via diffusion transformer.
\newblock \emph{arXiv preprint arXiv:2410.00086}, 2024.

\bibitem[He et~al.(2024)He, Ma, Huang, Huang, Gao, Wei, Dai, Han, and Liu]{he2024freeeditmaskfreereferencebasedimage}
Runze He, Kai Ma, Linjiang Huang, Shaofei Huang, Jialin Gao, Xiaoming Wei, Jiao Dai, Jizhong Han, and Si Liu.
\newblock Freeedit: Mask-free reference-based image editing with multi-modal instruction.
\newblock \emph{arXiv preprint arXiv:2409.18071}, 2024.

\bibitem[Hertz et~al.(2022)Hertz, Mokady, Tenenbaum, Aberman, Pritch, and Cohen-Or]{hertz2022prompt}
Amir Hertz, Ron Mokady, Jay Tenenbaum, Kfir Aberman, Yael Pritch, and Daniel Cohen-Or.
\newblock Prompt-to-prompt image editing with cross attention control.
\newblock \emph{arXiv preprint arXiv:2208.01626}, 2022.

\bibitem[Ho and Salimans(2022)]{cfg}
Jonathan Ho and Tim Salimans.
\newblock Classifier-free diffusion guidance.
\newblock \emph{arXiv preprint arXiv:2207.12598}, 2022.

\bibitem[Huang et~al.(2023)Huang, Tang, Dong, Lee, and Xu]{huang2023region}
Nisha Huang, Fan Tang, Weiming Dong, Tong-Yee Lee, and Changsheng Xu.
\newblock Region-aware diffusion for zero-shot text-driven image editing.
\newblock \emph{arXiv preprint arXiv:2302.11797}, 2023.

\bibitem[Huang et~al.(2024)Huang, Xie, Wang, Yuan, Cun, Ge, Zhou, Dong, Huang, Zhang, et~al.]{Smartedit}
Yuzhou Huang, Liangbin Xie, Xintao Wang, Ziyang Yuan, Xiaodong Cun, Yixiao Ge, Jiantao Zhou, Chao Dong, Rui Huang, Ruimao Zhang, et~al.
\newblock Smartedit: Exploring complex instruction-based image editing with multimodal large language models.
\newblock In \emph{CVPR}, 2024.

\bibitem[Labs(2024)]{flux2024}
Black~Forest Labs.
\newblock Flux.
\newblock \url{https://github.com/black-forest-labs/flux}, 2024.

\bibitem[Labs et~al.(2025)Labs, Batifol, Blattmann, Boesel, Consul, Diagne, Dockhorn, English, English, Esser, Kulal, Lacey, Levi, Li, Lorenz, Müller, Podell, Rombach, Saini, Sauer, and Smith]{labs2025flux1kontextflowmatching}
Black~Forest Labs, Stephen Batifol, Andreas Blattmann, Frederic Boesel, Saksham Consul, Cyril Diagne, Tim Dockhorn, Jack English, Zion English, Patrick Esser, Sumith Kulal, Kyle Lacey, Yam Levi, Cheng Li, Dominik Lorenz, Jonas Müller, Dustin Podell, Robin Rombach, Harry Saini, Axel Sauer, and Luke Smith.
\newblock Flux.1 kontext: Flow matching for in-context image generation and editing in latent space.
\newblock \emph{arXiv preprint arXiv:2506.15742}, 2025.

\bibitem[Li et~al.(2025)Li, Gu, Chen, Xing, Wen, Chen, and Zhu]{SuperEdit}
Ming Li, Xin Gu, Fan Chen, Xiaoying Xing, Longyin Wen, Chen Chen, and Sijie Zhu.
\newblock Superedit: Rectifying and facilitating supervision for instruction-based image editing.
\newblock \emph{arXiv preprint arXiv:2505.02370}, 2025.

\bibitem[Li et~al.(2023)Li, van~de Weijer, Hu, Khan, Hou, Wang, and Yang]{li2023stylediffusion}
Senmao Li, Joost van~de Weijer, Taihang Hu, Fahad~Shahbaz Khan, Qibin Hou, Yaxing Wang, and Jian Yang.
\newblock Stylediffusion: Prompt-embedding inversion for text-based editing.
\newblock \emph{arXiv preprint arXiv:2303.15649}, 2023.

\bibitem[Lipman et~al.(2023)Lipman, Chen, Ben-Hamu, Nickel, and Le]{lipman2023flowmatchinggenerativemodeling}
Yaron Lipman, Ricky T.~Q. Chen, Heli Ben-Hamu, Maximilian Nickel, and Matt Le.
\newblock Flow matching for generative modeling.
\newblock \emph{arXiv preprint arXiv:2210.02747}, 2023.

\bibitem[Liu et~al.(2023)Liu, Zeng, Ren, Li, Zhang, Yang, Li, Yang, Su, Zhu, et~al.]{liu2023grounding}
Shilong Liu, Zhaoyang Zeng, Tianhe Ren, Feng Li, Hao Zhang, Jie Yang, Chunyuan Li, Jianwei Yang, Hang Su, Jun Zhu, et~al.
\newblock Grounding dino: Marrying dino with grounded pre-training for open-set object detection.
\newblock \emph{arXiv preprint arXiv:2303.05499}, 2023.

\bibitem[Liu et~al.(2025)Liu, Han, Xing, Yin, Wang, Cheng, Liao, Wang, Fu, Han, Li, Peng, Sun, Wu, Cai, Ge, Ming, Xia, Zeng, Zhu, Jiao, Zhang, Yu, and Jiang]{step1x-edit}
Shiyu Liu, Yucheng Han, Peng Xing, Fukun Yin, Rui Wang, Wei Cheng, Jiaqi Liao, Yingming Wang, Honghao Fu, Chunrui Han, Guopeng Li, Yuang Peng, Quan Sun, Jingwei Wu, Yan Cai, Zheng Ge, Ranchen Ming, Lei Xia, Xianfang Zeng, Yibo Zhu, Binxing Jiao, Xiangyu Zhang, Gang Yu, and Daxin Jiang.
\newblock Step1x-edit: A practical framework for general image editing.
\newblock \emph{arXiv preprint arXiv:2504.17761}, 2025.

\bibitem[Liu et~al.(2022)Liu, Gong, and Liu]{liu2022flow}
Xingchao Liu, Chengyue Gong, and Qiang Liu.
\newblock Flow straight and fast: Learning to generate and transfer data with rectified flow.
\newblock \emph{arXiv preprint arXiv:2209.03003}, 2022.

\bibitem[Lugmayr et~al.(2022)Lugmayr, Danelljan, Romero, Yu, Timofte, and Gool]{lugmayr2022repaintinpaintingusingdenoising}
Andreas Lugmayr, Martin Danelljan, Andres Romero, Fisher Yu, Radu Timofte, and Luc~Van Gool.
\newblock Repaint: Inpainting using denoising diffusion probabilistic models.
\newblock \emph{arXiv preprint arXiv:2201.09865}, 2022.

\bibitem[Ma et~al.(2025)Ma, Wu, Sun, and Li]{hpsv3}
Yuhang Ma, Xiaoshi Wu, Keqiang Sun, and Hongsheng Li.
\newblock Hpsv3: Towards wide-spectrum human preference score.
\newblock \emph{arXiv preprint arXiv:2508.03789}, 2025.

\bibitem[Miyake et~al.(2024)Miyake, Iohara, Saito, and Tanaka]{miyake2024negativepromptinversionfastimage}
Daiki Miyake, Akihiro Iohara, Yu Saito, and Toshiyuki Tanaka.
\newblock Negative-prompt inversion: Fast image inversion for editing with text-guided diffusion models.
\newblock \emph{arXiv preprint arXiv:2305.16807}, 2024.

\bibitem[Mokady et~al.(2022)Mokady, Hertz, Aberman, Pritch, and Cohen-Or]{mokady2022null}
Ron Mokady, Amir Hertz, Kfir Aberman, Yael Pritch, and Daniel Cohen-Or.
\newblock Null-text inversion for editing real images using guided diffusion models.
\newblock \emph{arXiv preprint arXiv:2211.09794}, 2022.

\bibitem[{OpenAI et al.}(2024)]{openai2024gpt4ocard}
{OpenAI et al.}
\newblock Gpt-4o system card.
\newblock \emph{arXiv preprint arXiv:2410.21276}, 2024.

\bibitem[Oquab et~al.(2023)Oquab, Darcet, Moutakanni, Vo, Szafraniec, Khalidov, Fernandez, Haziza, Massa, El-Nouby, Howes, Huang, Xu, Sharma, Li, Galuba, Rabbat, Assran, Ballas, Synnaeve, Misra, Jegou, Mairal, Labatut, Joulin, and Bojanowski]{oquab2023dinov2}
Maxime Oquab, Timothée Darcet, Theo Moutakanni, Huy~V. Vo, Marc Szafraniec, Vasil Khalidov, Pierre Fernandez, Daniel Haziza, Francisco Massa, Alaaeldin El-Nouby, Russell Howes, Po-Yao Huang, Hu Xu, Vasu Sharma, Shang-Wen Li, Wojciech Galuba, Mike Rabbat, Mido Assran, Nicolas Ballas, Gabriel Synnaeve, Ishan Misra, Herve Jegou, Julien Mairal, Patrick Labatut, Armand Joulin, and Piotr Bojanowski.
\newblock Dinov2: Learning robust visual features without supervision.
\newblock \emph{arXiv preprint arXiv:2304.07193}, 2023.

\bibitem[Podell et~al.(2023)Podell, English, Lacey, Blattmann, Dockhorn, Müller, Penna, and Rombach]{sdxl}
Dustin Podell, Zion English, Kyle Lacey, Andreas Blattmann, Tim Dockhorn, Jonas Müller, Joe Penna, and Robin Rombach.
\newblock Sdxl: Improving latent diffusion models for high-resolution image synthesis.
\newblock \emph{arXiv preprint arXiv:2307.01952}, 2023.

\bibitem[Qu et~al.(2025)Qu, Cheng, Yang, Zhao, Lin, Shi, Li, Wang, Chua, and Jiang]{qu2025vincie}
Leigang Qu, Feng Cheng, Ziyan Yang, Qi Zhao, Shanchuan Lin, Yichun Shi, Yicong Li, Wenjie Wang, Tat-Seng Chua, and Lu Jiang.
\newblock Vincie: Unlocking in-context image editing from video.
\newblock \emph{arXiv preprint arXiv:2506.10941}, 2025.

\bibitem[Radford et~al.(2021)Radford, Kim, Hallacy, Ramesh, Goh, Agarwal, Sastry, Askell, Mishkin, Clark, Krueger, and Sutskever]{clip2021}
Alec Radford, Jong~Wook Kim, Chris Hallacy, Aditya Ramesh, Gabriel Goh, Sandhini Agarwal, Girish Sastry, Amanda Askell, Pamela Mishkin, Jack Clark, Gretchen Krueger, and Ilya Sutskever.
\newblock Learning transferable visual models from natural language supervision.
\newblock \emph{arXiv preprint arXiv:2103.00020}, 2021.

\bibitem[Ren et~al.(2025)Ren, Jiang, Zhang, Forchhammer, and S\"usstrunk]{fds}
Yufan Ren, Zicong Jiang, Tong Zhang, S{\o}ren Forchhammer, and Sabine S\"usstrunk.
\newblock Fds: Frequency-aware denoising score for text-guided latent diffusion image editing.
\newblock In \emph{CVPR}, 2025.

\bibitem[Rombach et~al.(2022)Rombach, Blattmann, Lorenz, Esser, and Ommer]{LDM}
Robin Rombach, Andreas Blattmann, Dominik Lorenz, Patrick Esser, and Björn Ommer.
\newblock High-resolution image synthesis with latent diffusion models.
\newblock \emph{arXiv preprint arXiv:2112.10752}, 2022.

\bibitem[Rout et~al.(2025)Rout, Chen, Ruiz, Caramanis, Shakkottai, and Chu]{rout2025semantic}
Litu Rout, Yujia Chen, Nataniel Ruiz, Constantine Caramanis, Sanjay Shakkottai, and Wen-Sheng Chu.
\newblock Semantic image inversion and editing using rectified stochastic differential equations.
\newblock In \emph{ICLR}, 2025.

\bibitem[Seedream et~al.(2025)Seedream, :, Chen, Gao, Gong, Guo, Guo, Guo, Hou, Huang, Huang, Jian, Kuang, Lai, Li, Li, Lian, Liao, Liu, Liu, Lu, Luo, Ou, Shi, Shi, Sun, Tian, Tian, Wang, Wang, Wang, Wang, Wu, Wu, Wu, Wu, Xia, Xiao, Xu, Yan, Yang, Yang, Zhai, Zhang, Zhang, Zhang, Zhang, Zhang, Zhao, Zhao, and Zhu]{seedream2025seedream40nextgenerationmultimodal}
Team Seedream, :, Yunpeng Chen, Yu Gao, Lixue Gong, Meng Guo, Qiushan Guo, Zhiyao Guo, Xiaoxia Hou, Weilin Huang, Yixuan Huang, Xiaowen Jian, Huafeng Kuang, Zhichao Lai, Fanshi Li, Liang Li, Xiaochen Lian, Chao Liao, Liyang Liu, Wei Liu, Yanzuo Lu, Zhengxiong Luo, Tongtong Ou, Guang Shi, Yichun Shi, Shiqi Sun, Yu Tian, Zhi Tian, Peng Wang, Rui Wang, Xun Wang, Ye Wang, Guofeng Wu, Jie Wu, Wenxu Wu, Yonghui Wu, Xin Xia, Xuefeng Xiao, Shuang Xu, Xin Yan, Ceyuan Yang, Jianchao Yang, Zhonghua Zhai, Chenlin Zhang, Heng Zhang, Qi Zhang, Xinyu Zhang, Yuwei Zhang, Shijia Zhao, Wenliang Zhao, and Wenjia Zhu.
\newblock Seedream 4.0: Toward next-generation multimodal image generation.
\newblock \emph{arXiv preprint arXiv:2509.20427}, 2025.

\bibitem[Sheynin et~al.(2024)Sheynin, Polyak, Singer, Kirstain, Zohar, Ashual, Parikh, and Taigman]{EmuEdit}
Shelly Sheynin, Adam Polyak, Uriel Singer, Yuval Kirstain, Amit Zohar, Oron Ashual, Devi Parikh, and Yaniv Taigman.
\newblock Emu edit: Precise image editing via recognition and generation tasks.
\newblock In \emph{CVPR}, 2024.

\bibitem[Simsar et~al.(2025)Simsar, Tonioni, Xian, Hofmann, and Tombari]{simsar2025uip2punsupervisedinstructionbasedimage}
Enis Simsar, Alessio Tonioni, Yongqin Xian, Thomas Hofmann, and Federico Tombari.
\newblock Uip2p: Unsupervised instruction-based image editing via edit reversibility constraint.
\newblock In \emph{ICCV}, 2025.

\bibitem[Song et~al.(2022)Song, Meng, and Ermon]{DDIM}
Jiaming Song, Chenlin Meng, and Stefano Ermon.
\newblock Denoising diffusion implicit models.
\newblock \emph{arXiv preprint arXiv:2010.02502}, 2022.

\bibitem[Sun et~al.(2023)Sun, Yang, Peng, Shen, Yang, Hu, Qiu, and Koike]{sun2023imagebrushlearningvisualincontext}
Yasheng Sun, Yifan Yang, Houwen Peng, Yifei Shen, Yuqing Yang, Han Hu, Lili Qiu, and Hideki Koike.
\newblock Imagebrush: Learning visual in-context instructions for exemplar-based image manipulation.
\newblock In \emph{NeurIPS}, 2023.

\bibitem[Tomasi and Manduchi(1998)]{bilateralfiltering}
C. Tomasi and R. Manduchi.
\newblock Bilateral filtering for gray and color images.
\newblock In \emph{Sixth International Conference on Computer Vision (IEEE Cat. No.98CH36271)}, 1998.

\bibitem[Tumanyan et~al.(2023)Tumanyan, Geyer, Bagon, and Dekel]{Tumanyan_2023_CVPR}
Narek Tumanyan, Michal Geyer, Shai Bagon, and Tali Dekel.
\newblock Plug-and-play diffusion features for text-driven image-to-image translation.
\newblock In \emph{CVPR}, 2023.

\bibitem[Wang et~al.(2024)Wang, Pu, Qi, Guo, Ma, Huang, Chen, Li, and Shan]{wang2024taming}
Jiangshan Wang, Junfu Pu, Zhongang Qi, Jiayi Guo, Yue Ma, Nisha Huang, Yuxin Chen, Xiu Li, and Ying Shan.
\newblock Taming rectified flow for inversion and editing.
\newblock \emph{arXiv preprint arXiv:2411.04746}, 2024.

\bibitem[Wang et~al.(2004)Wang, Bovik, Sheikh, and Simoncelli]{wang2004ssim}
Zhou Wang, A.C. Bovik, H.R. Sheikh, and E.P. Simoncelli.
\newblock Image quality assessment: from error visibility to structural similarity.
\newblock \emph{IEEE Transactions on Image Processing}, 2004.

\bibitem[Wu et~al.(2025)Wu, Li, Zhou, Lin, Gao, Yan, ming Yin, Bai, Xu, Chen, Chen, Tang, Zhang, Wang, Yang, Yu, Cheng, Liu, Li, Zhang, Meng, Wei, Ni, Chen, Cao, Peng, Qu, Wu, Wang, Yu, Wen, Feng, Xu, Wang, Zhang, Zhu, Wu, Cai, and Liu]{wu2025qwenimagetechnicalreport}
Chenfei Wu, Jiahao Li, Jingren Zhou, Junyang Lin, Kaiyuan Gao, Kun Yan, Sheng ming Yin, Shuai Bai, Xiao Xu, Yilei Chen, Yuxiang Chen, Zecheng Tang, Zekai Zhang, Zhengyi Wang, An Yang, Bowen Yu, Chen Cheng, Dayiheng Liu, Deqing Li, Hang Zhang, Hao Meng, Hu Wei, Jingyuan Ni, Kai Chen, Kuan Cao, Liang Peng, Lin Qu, Minggang Wu, Peng Wang, Shuting Yu, Tingkun Wen, Wensen Feng, Xiaoxiao Xu, Yi Wang, Yichang Zhang, Yongqiang Zhu, Yujia Wu, Yuxuan Cai, and Zenan Liu.
\newblock Qwen-image technical report.
\newblock \emph{arXiv preprint arXiv:2508.02324}, 2025.

\bibitem[Wu et~al.(2024)Wu, Fan, Qin, Gu, Zhao, and Chan]{wu2024freediffprogressivefrequencytruncation}
Wei Wu, Qingnan Fan, Shuai Qin, Hong Gu, Ruoyu Zhao, and Antoni~B. Chan.
\newblock Freediff: Progressive frequency truncation for image editing with diffusion models.
\newblock \emph{arXiv preprint arXiv:2404.11895}, 2024.

\bibitem[Xiao et~al.(2024)Xiao, Wang, Zhou, Yuan, Xing, Yan, Wang, Huang, and Liu]{OmniGen}
Shitao Xiao, Yueze Wang, Junjie Zhou, Huaying Yuan, Xingrun Xing, Ruiran Yan, Shuting Wang, Tiejun Huang, and Zheng Liu.
\newblock Omnigen: Unified image generation.
\newblock \emph{arXiv preprint arXiv:2409.11340}, 2024.

\bibitem[Zhang et~al.(2023)Zhang, Mo, Chen, Sun, and Su]{Zhang2023MagicBrush}
Kai Zhang, Lingbo Mo, Wenhu Chen, Huan Sun, and Yu Su.
\newblock Magicbrush: A manually annotated dataset for instruction-guided image editing.
\newblock In \emph{NeurIPS}, 2023.

\bibitem[Zhang et~al.(2018)Zhang, Isola, Efros, Shechtman, and Wang]{lpips2018}
Richard Zhang, Phillip Isola, Alexei~A. Efros, Eli Shechtman, and Oliver Wang.
\newblock The unreasonable effectiveness of deep features as a perceptual metric.
\newblock \emph{arXiv preprint arXiv:1801.03924}, 2018.

\bibitem[Zhang et~al.(2024)Zhang, Yang, Feng, Qin, Chen, Yu, Chen, Wang, Savarese, Ermon, Xiong, and Xu]{zhang2024hiveharnessinghumanfeedback}
Shu Zhang, Xinyi Yang, Yihao Feng, Can Qin, Chia-Chih Chen, Ning Yu, Zeyuan Chen, Huan Wang, Silvio Savarese, Stefano Ermon, Caiming Xiong, and Ran Xu.
\newblock Hive: Harnessing human feedback for instructional visual editing.
\newblock In \emph{CVPR}, 2024.

\bibitem[Zhang et~al.(2025)Zhang, Xie, Lu, Yang, and Yang]{zhang2025context}
Zechuan Zhang, Ji Xie, Yu Lu, Zongxin Yang, and Yi Yang.
\newblock In-context edit: Enabling instructional image editing with in-context generation in large scale diffusion transformer.
\newblock \emph{arXiv preprint arXiv:2504.20690}, 2025.

\bibitem[Zhou et~al.(2025)Zhou, Deng, He, Dong, and Tang]{MTC}
Zijun Zhou, Yingying Deng, Xiangyu He, Weiming Dong, and Fan Tang.
\newblock Multi-turn consistent image editing.
\newblock \emph{arXiv preprint arXiv:2505.04320}, 2025.

\end{thebibliography}
